\documentclass[10pt,twocolumn,letterpaper]{article}

\usepackage{cvpr} 
\usepackage{graphicx}
\usepackage{amsmath}
\usepackage{amssymb}
\usepackage{booktabs}
\usepackage[table]{xcolor}
\usepackage{colortbl}
\usepackage{enumitem}
\usepackage{pifont}
\usepackage{caption}
\usepackage{multirow}
\usepackage[pagebackref,breaklinks,colorlinks]{hyperref}
\usepackage[capitalize]{cleveref}
\crefname{section}{Sec.}{Secs.}
\Crefname{section}{Section}{Sections}
\Crefname{table}{Table}{Tables}
\crefname{table}{Tab.}{Tabs.}

\def\cvprPaperID{3351} 
\def\confName{CVPR}
\def\confYear{2024}

\begin{document}

\title{ViLa-MIL: Dual-scale Vision-Language Multiple Instance Learning for \\ Whole Slide Image Classification}

\author{Jiangbo Shi$^{1}$, Chen Li$^{1*}$, Tieliang Gong$^{1}$, Yefeng Zheng$^{2}$, Huazhu Fu$^{3}$\thanks{Co-corresponding authors.}\\
$^{1}$School of Computer Science and Technology, Xi'an Jiaotong University, Xi'an, China\\
$^{2}$Jarvis Research Center, Tencent YouTu Lab, Shenzhen, China\\
$^{3}$Institute of High Performance Computing, Agency for Science, Technology and Research, Singapore \\
{\tt\small shijiangbo@stu.xjtu.edu.cn, }{\tt\small cli@xjtu.edu.cn, }
{\tt\small hzfu@ieee.org}
}
\maketitle

\begin{abstract}
Multiple instance learning~(MIL)-based framework has become the mainstream for processing the whole slide image~(WSI) with giga-pixel size and hierarchical image context in digital pathology. 
However, these methods heavily depend on a substantial number of bag-level labels and solely learn from the original slides, which are easily affected by variations in data distribution. 
Recently, vision language model~(VLM)-based methods introduced the language prior by pre-training on large-scale pathological image-text pairs.
However, the previous text prompt lacks the consideration of pathological prior knowledge, therefore does not substantially boost the model's performance. 
Moreover, the collection of such pairs and the pre-training process are very time-consuming and source-intensive.
To solve the above problems, we propose a dual-scale vision-language multiple instance learning~(ViLa-MIL) framework for whole slide image classification. 
Specifically, we propose a dual-scale visual descriptive text prompt based on the frozen large language model~(LLM) to boost the performance of VLM effectively.
To transfer the VLM to process WSI efficiently, for the image branch, we propose a prototype-guided patch decoder to aggregate the patch features progressively by grouping similar patches into the same prototype;
for the text branch, we introduce a context-guided text decoder to enhance the text features by incorporating the multi-granular image contexts. 
Extensive studies on three multi-cancer and multi-center subtyping datasets demonstrate the superiority of ViLa-MIL. 
\end{abstract}

\begin{figure}
\centering
\includegraphics[width=\linewidth]{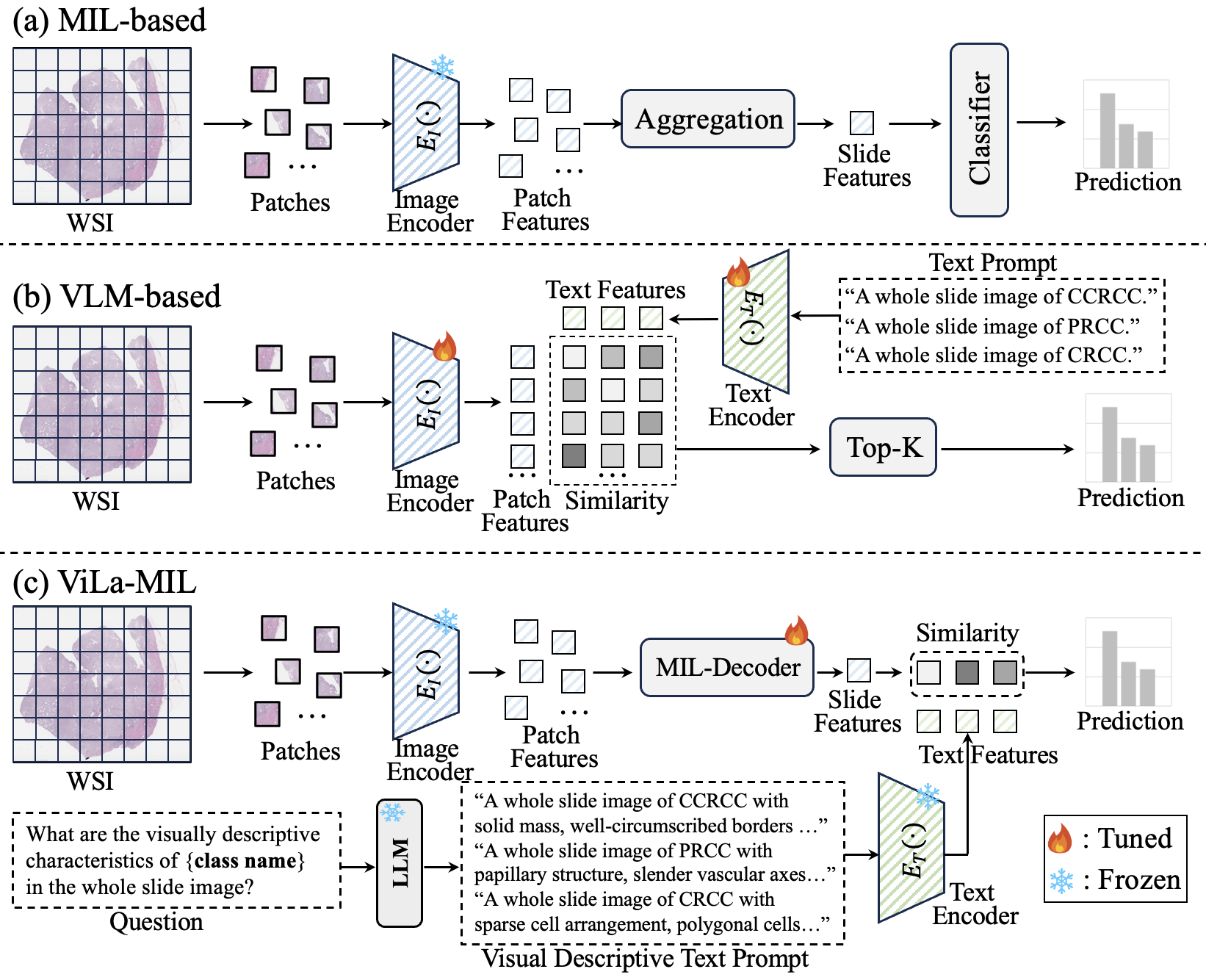}
\caption{Comparison of our ViLa-MIL with existing MIL- and VLM-based methods. (a) MIL-based methods design various aggregation functions to generate the slide-level features; (b) VLM-based methods calculate the similarity between patches and candidate text prompts, then utilize operator like top-K to obtain the slide-level prediction; (c) Our ViLa-MIL aligns the dual-scale slide-level features and the visual descriptive text prompt to obtain the slide prediction efficiently. Note that, for simplicity, only the single-scale data stream of ViLa-MIL is visualized.
}
\label{Motation}
\end{figure}

\section{Introduction}
Pathological examination serves as the gold standard for cancer diagnosis \cite{cui2021artificial}. With the fast development of digital scanning devices, traditional glass slides can be rapidly digitized into the whole slide image~(WSI) \cite{niazi2019digital}. 
WSI exhibits a hierarchical structure and huge size, which typically has about one billion pixels at its highest resolution~(0.25 $\mu$m/pixel) \cite{song2023artificial}. 
Therefore, obtaining a substantial dataset with accurate pixel-level annotations can often be challenging and unattainable. 
To alleviate this problem, various weakly supervised learning methods \cite{nakhli2023sparse, tang2023multiple, lin2023interventional, li2023task, ryu2023ocelot} have been proposed. 
Among them, multiple instance learning~(MIL) \cite{ilse2018attention, li2021dual, lu2021data, H2MIL} has gained significant popularity for tackling WSI classification tasks. 

MIL utilizes small patches~(\textit{i.e.}, instances) to generate the slide-level~(\textit{i.e.}, bag-level) representation \cite{maron1997framework} for analysis. 
As shown in Figure \ref{Motation}(a), MIL-based methods \cite{ilse2018attention, li2021dual, shao2021transmil} usually follow a three-step pipeline: patch cropping, feature extraction using a pretrained encoder, and slide-level feature aggregation for WSI classification. 
They \cite{nakhli2023sparse, tang2023multiple, lin2023interventional, li2023task, ryu2023ocelot} have achieved significant success in various pathological diagnostic tasks like cancer subtyping \cite{li2023task, chan2023histopathology, 10244116}, staging \cite{shi2023structure, H2MIL}, and tissue segmentation \cite{xu2019camel, li2023weakly}. 
However, training these MIL-based models still heavily relies on a large number of slides with bag-level labels which are often unreachable for rare diseases.
Moreover, these models only learn from the original slide, making them vulnerable to variations in data distribution and leading to suboptimal generalization performance.
Recently, a series of emerging works based on visual language model~(VLM) such as CLIP \cite{radford2021learning} and BLIP \cite{li2022blip} introduced the language information, bringing new advancements to the fields of natural image classification \cite{yu2023task, guo2023texts}, segmentation \cite{lin2023clip, rao2022denseclip}, object detection \cite{wu2023cora, zhong2022regionclip}, etc. 
In the field of pathology, several VLM-based methods like MI-Zero \cite{lu2023visual}, BiomedCLIP \cite{zhang2023large}, PLIP \cite{huang2023visual}, and QUILT \cite{ikezogwo2023quilt} have also been proposed and achieved promising results in diverse pathological diagnostic tasks and datasets. 
As shown in Figure \ref{Motation}(b), following the VLM-based framework, these methods \cite{lu2023visual, huang2023visual, lu2023towards, ikezogwo2023quilt} adopt the two-tower encoders~(\textit{i.e.}, image and text encoders), which are pre-trained on a large number of patch-text pairs collected from the Internet like Twitter or public educational resources \cite{gamper2021multiple, ikezogwo2023quilt}. 
Although these models have demonstrated great classification performance in various tasks, they are still subject to the following limitations:

Firstly, the text prompt does not provide effective guidance for identifying ambiguous categories, as it lacks the consideration of pathological prior knowledge. 
During inference, they exploit lots of handcrafted text prompt templates~(\textit{e.g.}, ``An H\&E image of a \{class name\}.") by filling in each template with the candidate category names. However, such a class-name-replacement text template only provides the class name as discriminative information, which may be insufficient for accurate classification. 
Introducing visual descriptive texts can help the model focus on diagnosis-related features and enhance its discriminative ability in classes with subtle differences.
For example, both breast mucinous and tubular carcinoma manifest similar glandular structures. 
However, each subtype possesses unique features that can aid in their diagnosis. 
Specifically, mucinous carcinoma is characterized by mucin production, while tubular carcinoma forms small tubular structures.

Secondly, transferring the VLM-based model to the field of pathology in a parameter-efficient way is challenging.
The current VLM-based methods \cite{lu2023visual, huang2023visual, lu2023towards} in pathology heavily rely on a large number of pathological image-text pairs, which are very time-consuming and laborious to collect. Moreover, the pre-training process requires substantial computational resources and time. 
VLM like CLIP \cite{radford2021learning} has been pre-trained on over 400 million image-text pairs, including medical images. 
It may be feasible to directly apply the VLM to the field of pathology.
However, given the huge size and hierarchical context of WSI, how to efficiently transfer knowledge in the VLM to process WSI is still challenging.

To address the above limitations, in this work, we propose a dual-scale vision-language multiple instance learning framework~(named as ViLa-MIL) for whole slide image classification.
As shown in Figure \ref{Motation}(c), the core idea of ViLa-MIL is to efficiently transfer the VLM to the pathology domain by better harnessing the prior knowledge of the routine pathological diagnosis. \textbf{\textit{Firstly, we propose a dual-scale visual descriptive text prompt to boost the performance of VLM effectively}}.
Inspired by the routine diagnosis of pathologists, we construct our dual-scale visual descriptive text prompt, which corresponds to WSIs at different resolutions, based on the frozen large language model~(LLM) \cite{brown2020language}. 
Specifically, the low-scale visual descriptive text prompt mainly focuses on the global tumor structure~(\textit{e.g.}, glandular or papillary structure) presented in a WSI of low magnification~(\textit{i.e.}, low-resolution), while the high-scale visual descriptive text prompt pays more attention to local finer details~(\textit{e.g.}, clear cytoplasm and round nuclei) on a high-resolution WSI. 
In this way, the text prompt can guide the VLM to uncover more discriminative diagnostic-related morphological patterns from the WSI. 

\textbf{\textit{Secondly, we propose two lightweight and trainable decoders for image and text branches, respectively, to adapt the VLM for processing the WSI efficiently}}.
Specifically, for the image branch, to aggregate all the patch features, we propose a prototype-guided patch decoder. 
For a WSI of a specific magnification, a set of learnable prototype vectors is introduced to progressively guide the fusion process of patch features by grouping similar patch features into the same prototype. 
In this way, each prototype captures more global contextual information for final similarity calculation.
For the text branch, we propose a context-guided text decoder. By introducing the multi-granular image context~(\textit{i.e.}, local patch features and global prototype features) as the guidance,  the pre-trained knowledge in the VLM text encoder can be better utilized. 

\textbf{\textit{Thirdly, ViLa-MIL achieves the best results compared with the current state-of-the-art MIL-based methods under the few-shot settings.}}
In the experiment, three multi-caner and multi-center datasets are used. Specifically, ViLa-MIL outperforms the state-of-the-art methods by 1.7-7.2\%, and 2.1-7.3\% in area under the curve~(AUC) and F1 score, respectively. 
Additionally, a series of ablation studies and parameter sensitivity analyses demonstrate the effectiveness of each module within ViLa-MIL.

\section{Related Works}
\noindent\textbf{Multiple Instance Learning in WSI.}
Recently, MIL-based methods \cite{ilse2018attention, li2021dual, H2MIL, shao2021transmil, lu2021data, zheng2021deep, zhang2022dtfd, lu2023visual, huang2023conslide} have become the mainstream for processing the whole slide image in the field of computational pathology. The whole process mainly includes three steps: 1) a series of patches are cropped from the original WSI; 2) a pre-trained encoder is utilized to extract the patch features; 3) the patch features are aggregated to generate the final slide features. Conventional handcrafted patch feature aggregators include non-parametric max and mean pooling operators. Later, ABMIL \cite{ilse2018attention} proposes an attention-based aggregation function by learning the weight for each patch through a parameterized neural network. CLAM \cite{lu2021data} utilizes a pre-trained image encoder for patch feature extraction and proposes a multi-branch pooling operator trained for weakly-supervised WSI classification tasks. 
Based on CLAM, a series of methods \cite{shao2021transmil, zheng2021deep, zhang2022dtfd, lu2023visual, li2021dual} have been proposed to explore how to aggregate the patch features effectively. 
For example, DSMIL \cite{li2021dual} utilizes the multi-scale patch features as the input and aggregates it in a dual-stream architecture. 
TransMIL \cite{shao2021transmil} explores both morphological and spatial relations through several self-attention layers. 
GTMIL \cite{zheng2021deep} adopts a graph-based representation and vision Transformer for WSI. 
DTMIL \cite{zhang2022dtfd} utilizes a double-tier MIL framework by introducing the concept of pseudo-bags.
IBMIL \cite{lin2023interventional} achieves deconfounded bag-level prediction based on backdoor adjustment.
Although these methods have achieved great success in many pathological diagnostic tasks, they rely solely on bag-level labels for training, thus requiring the model to learn discriminative patterns from a large quantity of WSIs, without fully utilizing pathological prior knowledge as the guideline. 
In this work, our ViLa-MIL introduces the dual-scale visual descriptive text prompt as the language prior to guide the training of the model effectively. 

\noindent\textbf{Visual Language Models in WSI.}
The vision language models, like CLIP \cite{radford2021learning} and FLIP \cite{li2023scaling}, have shown remarkable performance in a wide variety of visual recognition tasks \cite{li2022blip, yu2023task, guo2023texts, khattak2023maple, lin2023clip, han20232vpt}. 
CLIP-based methods utilize a dual-tower structure, including an image encoder and a text encoder, which are pre-trained on a large number of web-source data by aligning the image-text pairs in the same feature space. 
Leveraging a class-name-replacement~(\textit{e.g.}, ``an image of \{class name\}."), CLIP classifies an image by matching it to the candidate text prompt of the highest similarity. 
In the field of pathology, several works~(\textit{e.g.}, MI-Zero \cite{lu2023visual} and PLIP \cite{huang2023visual}) have also been proposed by utilizing the CLIP model. 
For example, MI-Zero \cite{lu2023visual} aligns image and text models on pathological image patches by pre-training on 33k image-caption pairs. 
Later, PLIP \cite{huang2023visual} builds a large dataset of 208,414 pathological images paired with language descriptions and utilizes this dataset to fine-tune the CLIP model to the pathological field. 
QUILT \cite{ikezogwo2023quilt} also fine-tunes the CLIP model by creating over one million paired image-text samples based on YouTube and some public education resources.
Although these works \cite{lu2021data, huang2023visual, zhang2023text} have demonstrated significant classification performance and transferability in a new dataset, collecting a large number of image-text pairs is extremely time-consuming and labor-intensive. 
In the natural image field, several methods, such as CoOp \cite{zhou2022learning}, CLIP-Adapter \cite{gao2023clip} and TaskRes \cite{yu2023task}, have been proposed to transfer the CLIP model in a parameter-efficient fine-tuning way. However, these methods \cite{zhou2022learning, gao2023clip, yu2023task, guo2023texts} fail to take into account the hierarchical structure and large size of WSI, as well as the integration of the pathological knowledge.  
Recently, TOP \cite{qu2024rise} proposes a two-level prompt learning MIL framework, incorporating language prior knowledge. However, it still only utilizes the single-scale patches and the text prompt features are solely optimized based on the labels.
In this work, we utilize the frozen large language model to generate the dual-scale visual descriptive text prompt, which aids the model in transferring to the target dataset with the guidance of limited labeled data. Moreover, two lightweight and trainable decoders are proposed, with one for the image branch and the other for text, to transfer the VLM model to the field of pathology efficiently.

\begin{figure*}
\centering
\includegraphics[width=\linewidth]{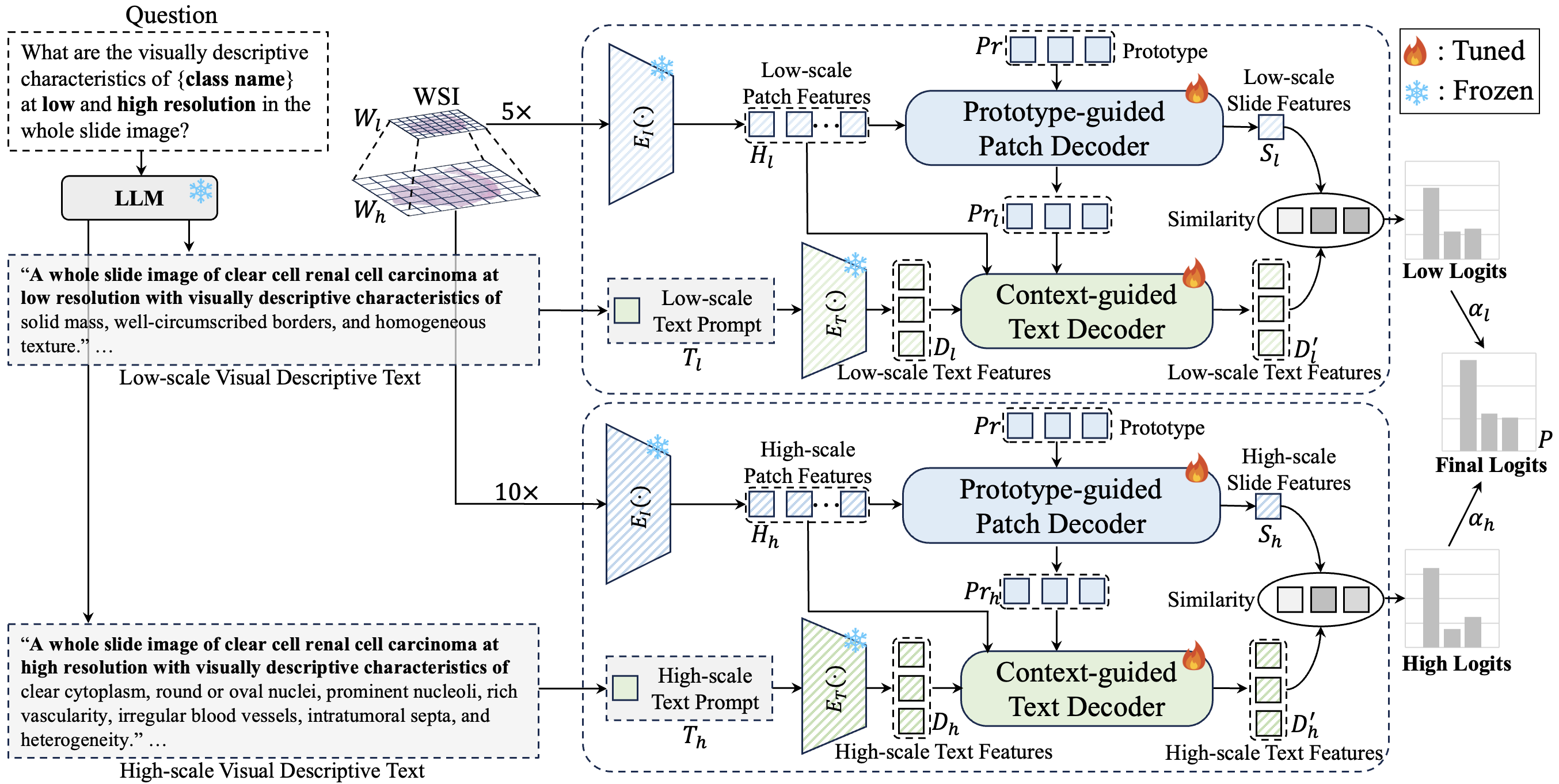}
\caption{
Pipeline of the proposed ViLa-MIL framework. 
The input of ViLa-MIL is a Question and WSI. 
The question is passed through a frozen large language model~(LLM) to generate the dual-scale visual descriptive text prompt. 
The prototype-guided patch decoder is introduced to progressively fuse the patch features into the slide features. 
The context-guided text decoder is introduced to refine the text features further by utilizing the multi-granular image contexts. 
}
\label{framework}
\vspace{-0.5cm}
\end{figure*}

\section{Methodology}
\subsection{Revisiting CLIP}
Many VLMs \cite{radford2021learning, li2022blip, li2023scaling} have shown great performance in various image classification tasks.
In this work, we select the CLIP model \cite{radford2021learning} as the baseline.
Specifically, CLIP adopts a two-tower model structure including an image encoder and a text encoder. For each image-text pair, the image is fed into the image encoder $E_I(\cdot)$ to generate its visual features and the text features are generated by passing the text to the text encoder $E_T(\cdot)$. The contrastive loss pulls close the image and its corresponding descriptions in a unified embedding space and pushes away the features of non-paired images and texts. 
During inference, CLIP classifies an image into $C$ possible categories by calculating the similarity between the image features and all the candidate text features.
Formally, let $I$ denote the image and $T$ denote all the text templates~(\textit{e.g.}, ``an image of a \{class name\}"). Given the image features $x$ and text features $\{t_i\}_{i=1}^{C}$, the predicted probability for class $i$ is defined as 
\begin{equation}
\label{eq1}
\setlength\abovedisplayskip{3pt}
\setlength\belowdisplayskip{3pt}
p(y=i|x) = \frac{\mathrm{exp}(\mathrm{cos}(x, t_i))}{\sum_{j=1}^{C}\mathrm{exp}(\mathrm{cos}(x, t_j))},
\end{equation}
where $\mathrm{cos}(\cdot, \cdot)$ denotes the cosine similarity. 

\subsection{Dual-scale Visual Descriptive Text Prompt}
To boost the CLIP model effectively, one of the most important questions is how to construct a text prompt to describe the whole slide image.
During the routine slide reviewing process, pathologists typically start by examining the overall structure of tissues at a relatively low magnification. Subsequently, they zoom in to a higher magnification to check finer details or textures, such as the size and shape of the nuclei. 
    Combining this diagnostic prior with the multi-scale characteristics of WSI, we propose our dual-scale visual descriptive text prompt to guide the CLIP model for WSI classification, as shown in Figure \ref{framework}. 

The specific dual-scale visual descriptive text prompt generation process is as follows. A text question prompt is first designed for the frozen large language model:
\textit{``What are the visually descriptive characteristics of \{class name\} at low and high resolution in the whole slide image?"}
Taking the renal cancer subtyping task as an example, by replacing the class name with the specific categories~(\textit{i.e.}, \textit{clear cell~(CCRCC), papillary~(PRCC), and chromophobe renal cell carcinoma~(CRCC)}), the corresponding descriptive text prompts can be generated automatically by the LLM. For example, the low-scale text prompt of PRCC mainly focuses on tissue structures present in a low-resolution WSI, like \textit{``the papillary growth pattern"} and \textit{``well-circumscribed borders"}. The high-scale text prompt of PRCC mainly focuses on the details in a high-resolution WSI, like \textit{``nuclei arranged in layers"} and \textit{``heterogeneous cytoplasm"}.
These dual-scale visual descriptive texts are consistent with the daily practice of pathologists, which can help the model distinguish subtle and fine-grained pathological morphological features and improve the model's classification performance.
To better transfer the CLIP knowledge to the pathological field, inspired by the CoOp \cite{zhou2022learning}, we also add $M$ learnable vectors into the dual-scale text prompts as the supplementary prompt.
Formally, the dual-scale visual descriptive text prompt is denoted as $T_l$ and $T_h$ for low and high scales, respectively, as follows:
\begin{equation}
\setlength\abovedisplayskip{3pt}
\setlength\belowdisplayskip{3pt}
\scalebox{0.95}{
    $T_l = [V_l]_1[V_l]_2...[V_l]_M[{Low\text{-}scale\ Text\ Prompt}],$
}
\end{equation}
\begin{equation}
\setlength\abovedisplayskip{3pt}
\setlength\belowdisplayskip{3pt}
\scalebox{0.95}{
    $T_h = [V_h]_1[V_h]_2...[V_h]_M[{High\text{-}scale\ Text\ Prompt}],$
}
\end{equation}
where $[V_*]_i, i \in \{1, ...,M\}$ are the learnable vectors.

\subsection{Prototype-guided Patch Decoder}
\begin{figure}
\centering
\includegraphics[width=\linewidth]{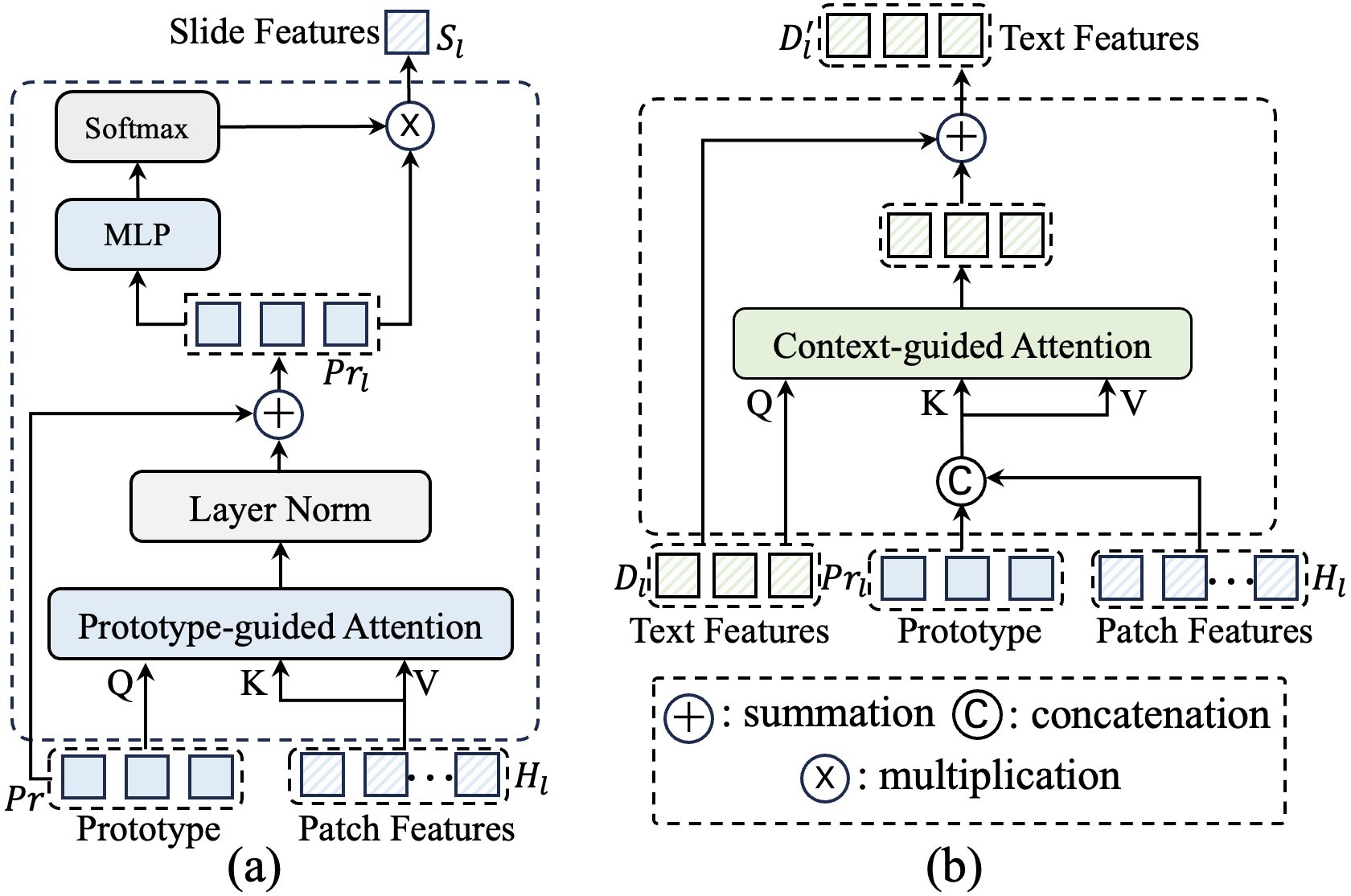}
\caption{(a) Prototype-guided patch decoder; (b) Context-guided text decoder.}
\label{decoder}
\vspace{-0.6cm}
\end{figure}

To apply the CLIP model to process WSI efficiently, another important question is how to aggregate a large number of patch features.
In the multiple instance learning~(MIL) framework, a WSI $W = \{W_l, W_h\}$~(where $W_l$ and $W_h$ represent the original slides at low and high magnification, respectively.) with hierarchical structure is taken as a bag containing multiple instances as $I = \{I_l \in \mathbb{R}^{N_l{\times}N_0{\times}N_0{\times}3}, I_h \in \mathbb{R}^{N_h{\times}N_0{\times}N_0{\times}3}\}$
~(where $N_l$ and $N_h$ denote the number of low- and high-resolution patches, respectively, and $N_0$ is the patch size). 
Following the current embedding-based MIL methods \cite{ilse2018attention, lu2021data}, a non-overlapping sliding window method is utilized to crop patches $I$ from the WSI. 
Then, the frozen CLIP image encoder $E_I(\cdot)$ is utilized to map the patches $I$ into a feature vector $H = \{H_l \in \mathbb{R}^{N_l{\times}d}, H_h \in \mathbb{R}^{N_h{\times}d}\}$~(where $d$ is the feature dimension). 
To efficiently aggregate a large number of patch features into the final slide features for the calculation of similarity, we design a prototype-guided patch decoder. 
As shown in Figure \ref{decoder}(a), we randomly initialize a set of learnable prototype features $Pr \in \mathbb{R}^{N_p{\times}d}$ to guide the progressive fusion of patch features $H_l$.\footnote{Without loss of generality, we take the low-scale~(\textit{i.e.}, low-resolution) patch feature fusion process as an example.} Specifically, the defined prototypes $Pr$ and low-scale patch features $H_l$ are first fed into a prototype-guided attention layer, which is implemented as a cross-attention layer. Formally, we take the prototypes $Pr$ as the query $Q_l$, all the low-scale patch features $H_l$ as the key $K_l$ and value $V_l$. The prototype-guided attention between the prototypes and patch features is as follows:
\begin{equation}
\setlength\abovedisplayskip{3pt}
\setlength\belowdisplayskip{3pt}
    Pr_{l} = \mathrm{Norm}(\mathrm{Softmax}(\frac{Q_lK_l^{\top}}{\sqrt{d}})V_l) + Pr,
\end{equation}
where $\mathrm{Norm}(\cdot)$ denotes the layer normalization operator and $\mathrm{Softmax(\cdot)}$ is the activation function. 
By introducing the guidance of learnable prototypes, the patches with high semantic similarity will be grouped into the same prototype. Compared with the local patches with the limited receptive field, each prototype captures more global context information. Finally, to obtain the final slide-level features $S_{l}$, we utilize the attention-based feature fusion method:
\begin{equation}
\setlength\abovedisplayskip{3pt}
\setlength\belowdisplayskip{3pt}
    Pr_{l, i}^{\prime} = \mathbf{W_a}Pr_{l, i},
\end{equation}
\begin{equation}
\setlength\abovedisplayskip{3pt}
\setlength\belowdisplayskip{3pt}
    A_{l, i} = \frac{\mathrm{exp}\{\mathbf{W_b^\top}(\mathrm{tanh}(\mathbf{W_v}Pr_{l, i}^{\prime^\top}))\}}{\sum_{j=\text{1}}^{N_p}{\mathrm{exp}\{\mathbf{W_b^\top}(\mathrm{tanh}(\mathbf{W_v}Pr_{l, j}^{\prime^\top}))\}}},
\end{equation}
\begin{equation}
\setlength\abovedisplayskip{3pt}
\setlength\belowdisplayskip{3pt}
    S_{l} = \mathbf{W_c}\sum_{i=1}^{N_p}{{A}_{l,i}Pr_{l, i}^{\prime}},
\end{equation}
where $\mathbf{W_a}$, $\mathbf{W_c}$, $\mathbf{W_v}$ are trainable weight matrices with the same dimension of $d \times d$; $\mathbf{W_b} \in \mathbb{R}^{d \times 1}$ is a trainable weight matrix; $\mathrm{tanh}(\cdot)$ is the activation function; and $A_l$ is the attention map that assigns the weight of each prototype in the final slide-level representation $S_l$. 
The high-scale slide-level representation $S_h$ can be generated in a similar way.

\begin{table*}[]
\centering
\scalebox{0.8}{
\begin{tabular}{@{}l|lll|lll|lll@{}}
\toprule
Dataset      & \multicolumn{3}{c|}{TIHD-RCC} & \multicolumn{3}{c|}{TCGA-RCC} & \multicolumn{3}{c}{TCGA-Lung} \\ \midrule
Metric &
  \multicolumn{1}{c}{AUC} &
  \multicolumn{1}{c}{F1} &
  \multicolumn{1}{c|}{ACC} &
  \multicolumn{1}{c}{AUC} &
  \multicolumn{1}{c}{F1} &
  \multicolumn{1}{c|}{ACC} &
  \multicolumn{1}{c}{AUC} &
  \multicolumn{1}{c}{F1} &
  \multicolumn{1}{c}{ACC} \\ \midrule
Max-pooling &
  52.5 $\pm$ 4.9 &
  32.2 $\pm$ 3.5 &
  37.5 $\pm$ 2.3 &
  67.4 $\pm$ 4.9 &
  46.7 $\pm$ 11.6 &
  54.1 $\pm$ 4.8 &
  53.0 $\pm$ 6.0 &
  45.8 $\pm$ 8.9 &
  53.3 $\pm$ 3.4 \\
Mean-pooling &
  67.2 $\pm$ 4.6 &
  42.4 $\pm$ 10.1 &
  47.9 $\pm$ 3.9 &
  83.3 $\pm$ 6.0 &
  60.9 $\pm$ 8.5 &
  62.3 $\pm$ 7.4 &
  67.4 $\pm$ 7.2 &
  61.1 $\pm$ 5.5 &
  61.9 $\pm$ 5.5 \\
ABMIL \cite{ilse2018attention} &
  65.6 $\pm$ 9.6 &
  46.9 $\pm$ 7.1 &
  47.9 $\pm$ 6.7 &
  83.6 $\pm$ 3.1 &
  64.4 $\pm$ 4.2 &
  65.7 $\pm$ 4.7 &
  60.5 $\pm$ 15.9 &
  56.8 $\pm$ 11.8 &
  61.2 $\pm$ 6.1 \\
CLAM-SB \cite{lu2021data} &
  68.9 $\pm$ 14.4 &
  50.5 $\pm$ 16.3 &
  53.1 $\pm$ 13.8 &
  90.1 $\pm$ 2.2 &
  75.3 $\pm$ 7.4 &
  77.6 $\pm$ 7.0 &
  66.7 $\pm$ 13.6 &
  59.9 $\pm$ 13.8 &
  64.0 $\pm$ 7.7 \\
CLAM-MB \cite{lu2021data} &
  71.8 $\pm$ 9.8 &
  54.8 $\pm$ 9.5 &
  55.4 $\pm$ 10.2 &
  90.9 $\pm$ 4.1 &
  76.2 $\pm$ 4.4 &
  78.6 $\pm$ 4.9 &
  68.8 $\pm$ 12.5 &
  60.3 $\pm$ 11.1 &
  63.0 $\pm$ 9.3 \\
TransMIL \cite{shao2021transmil} &
  71.2 $\pm$ 4.2 &
  53.9 $\pm$ 3.9 &
  54.9 $\pm$ 4.1 &
  89.4 $\pm$ 5.6 &
  73.0 $\pm$ 7.8 &
  75.3 $\pm$ 7.2 &
  64.2 $\pm$ 8.5 &
  57.5 $\pm$ 6.4 &
  59.7 $\pm$ 5.4 \\
DSMIL \cite{li2021dual} &
  69.9 $\pm$ 9.4 &
  49.9 $\pm$ 6.9 &
  50.0 $\pm$ 7.4 &
  87.6 $\pm$ 4.5 &
  71.5 $\pm$ 6.6 &
  72.8 $\pm$ 6.4 &
  67.9 $\pm$ 8.0 &
  61.0 $\pm$ 7.0 &
  61.3 $\pm$ 7.0 \\
GTMIL \cite{zheng2021deep} &
  77.1 $\pm$ 3.5 &
  61.4 $\pm$ 4.3 &
  62.2 $\pm$ 4.3 &
  88.1 $\pm$ 13.3 &
  71.1 $\pm$ 15.7 &
  76.1 $\pm$ 12.9 &
  66.0 $\pm$ 15.3 &
  61.1 $\pm$ 12.3 &
  63.8 $\pm$ 9.9 \\
DTMIL \cite{zhang2022dtfd} &
  70.3 $\pm$ 10.3 &
  55.6 $\pm$ 9.6 &
  55.8 $\pm$ 9.9 &
  90.0 $\pm$ 4.6 &
  74.4 $\pm$ 5.3 &
  76.8 $\pm$ 5.2 &
  67.5 $\pm$ 10.3 &
  57.3 $\pm$ 11.3 &
  66.6 $\pm$ 7.5$^*$ \\
IBMIL \cite{lin2023interventional} &
  71.5 $\pm$ 6.3 &
  57.2 $\pm$ 7.6 &
  56.4 $\pm$ 4.8 &
  90.5 $\pm$ 4.1 &
  75.1 $\pm$ 5.2 &
  77.2 $\pm$ 4.2 &
  69.2 $\pm$ 7.4 &
  57.4 $\pm$ 8.3 &
  66.9 $\pm$ 6.5$^*$ \\
\rowcolor{gray!20} ViLa-MIL Low &
  79.5 $\pm$ 3.6 &
  61.2 $\pm$ 7.1 &
  63.1 $\pm$ 3.0 &
  90.9 $\pm$ 2.9 &
  77.3 $\pm$ 4.0 &
  79.5 $\pm$ 3.7 &
  71.9 $\pm$ 6.2 &
  64.1 $\pm$ 7.8 &
  65.8 $\pm$ 5.8 \\
  \rowcolor{gray!20} ViLa-MIL High &
  \underline{83.0 $\pm$ 5.6} &
  \underline{65.3 $\pm$ 6.4} &
  \underline{66.2 $\pm$ 6.8} &
  \underline{92.0 $\pm$ 1.3} &
  \underline{77.8 $\pm$ 3.7} &
  \underline{80.0 $\pm$ 3.0} &
  \underline{72.3 $\pm$ 7.6} &
  \underline{66.6 $\pm$ 6.5} &
  \underline{66.9 $\pm$ 6.3} \\ 
  \rowcolor{gray!20} \textbf{ViLa-MIL} &
  \textbf{84.3 $\pm$ 4.6} &
  \textbf{68.7 $\pm$ 7.3} &
  \textbf{68.8 $\pm$ 7.3} &
  \textbf{92.6 $\pm$ 3.0} &
  \textbf{78.3 $\pm$ 6.9} &
  \textbf{80.3 $\pm$ 6.2} &
  \textbf{74.7 $\pm$ 3.5} &
  \textbf{67.0 $\pm$ 4.9} &
  \textbf{67.7 $\pm$ 4.4} \\ \bottomrule
\end{tabular}}
\caption{Results~(presented in \%) on three datasets under the 16-shot setting. The top result is in bold, the second best result is underlined, and the comparable performance is denoted by superscript * based on a paired t-test~(p-value $>$ 0.05).}
\label{exp1}
\vspace{-0.5cm}
\end{table*}

\subsection{Context-guided Text Decoder}
For the text branch, the low-scale visual descriptive text prompt $T_l$ is first fed into the frozen text encoder $E_T(\cdot)$ to generate the low-scale text features $D_l$. We leverage the multi-granular visual contextual information~(\textit{i.e.}, local- patch and global-prototype features) to further refine the text features. By bridging the gap between the two modalities, the model can achieve better alignment of images and texts. As shown in Figure \ref{decoder}(b), a context-guided attention layer is adopted to achieve this refinement process. Specifically, 
the low-scale prototype features $Pr_l$ and patch features $H_l$ are first concatenated together as the key $K_{t, l}$ and value $V_{t, l}$. The low-scale text features $D_{l}$ are taken as the query $Q_{t, l}$. The context-guided attention is implemented with a cross-attention layer in the following manner:
\begin{equation}
\setlength\abovedisplayskip{3pt}
\setlength\belowdisplayskip{3pt}
    D_l^{\prime} = \mathrm{Softmax}(\frac{Q_{t, l}K_{t, l}^{\top}}{\sqrt{d}})V_{t, l} + D_l.
\end{equation}
The updated high-scale text features $D_h^{\prime}$ of the text prompt $T_h$ can be generated in a similar way.

\subsection{Training Strategy}
After obtaining the slide-level image features $S_l, S_h$ and refined text features $D_l^{\prime}, D_h^{\prime}$ for low and high scales, respectively, we can calculate the similarity between slide and text features of category $i$ based on Eq. (\ref{eq1}) as follows:
\begin{equation}
\setlength\abovedisplayskip{3pt}
\setlength\belowdisplayskip{3pt}
        P_i = \sum_{k=\{l,h\}}{\alpha_k*\frac{\mathrm{exp}(\mathrm{cos}(S_{k, i}, D_{k, i}^{\prime}))}{\sum_{j=1}^{C}\mathrm{exp}(\mathrm{cos}(S_{k, i}, D_{k, j}^{\prime}))}},
\end{equation}
where $\alpha_l$ and $\alpha_h$ control the weight of each-scale similarity.
Finally, the whole model is trained end-to-end, and the cross-entropy loss is formally defined as $L = \mathrm{CE}(P, GT)$,
where $\mathrm{CE(\cdot, \cdot)}$ denotes the cross entropy loss and $GT$ represents the slide-level labels.

\section{Experiments}
\subsection{Settings}
\noindent\textbf{Datasets}. 
We evaluate our ViLa-MIL on three real-world WSI subtyping datasets, namely TIHD-RCC, TCGA-RCC, and TCGA-Lung,\footnote{https://portal.gdc.cancer.gov.} under the few-shot setting. Note that TIHD-RCC is the in-house dataset collected by our team.  
Each dataset is split into the training, validation, and test sets with a ratio of 4:3:3.
To build the few-shot setting, we then randomly select 16 samples for each category in the training set as training samples.

\noindent\textbf{Implementation Details}. The original WSI is processed by Otsu's binarization algorithm first to filter out all the non-tissue regions. 
The stain of patches is pre-processed with the z-score normalization. 
The cropped patch size is $256 \times 256$ pixels~(\textit{i.e.}, $N_0=256$). 
GPT-3.5 is taken as the frozen large language model~(LLM). ResNet-50 is taken as the image encoder $E_I(\cdot)$, and the corresponding CLIP Transformer is used as the text encoder $E_T(\cdot)$.
The feature dimension $d$ is 1024. The number of prototype vectors $Pr$ is 16 and the number of context tokens $M$ is 16. The regularization parameters $\alpha_l$ and $\alpha_h$ are both 1.
The code and datasets are available at \url{https://github.com/Jiangbo-Shi/ViLa-MIL}. 

\noindent\textbf{Evaluation Metrics}. The area under the curve~(AUC) score, F1 score~(F1), and accuracy~(ACC) are utilized as the evaluation metrics in the experiment. 
We conduct the experiment for each method five times.
Specifically, for each run, we randomly split the dataset according to the prescribed proportion and then select 16 samples for each category in the training set to train the model.
The mean and standard deviation values are calculated based on the five results. 
The paired t-test as a statistical method is also adopted to determine whether the two paired results are comparable. 
When the p-value is larger than 0.05, these two results are statistically comparable.

\begin{figure}
\centering
\includegraphics[width=\linewidth]{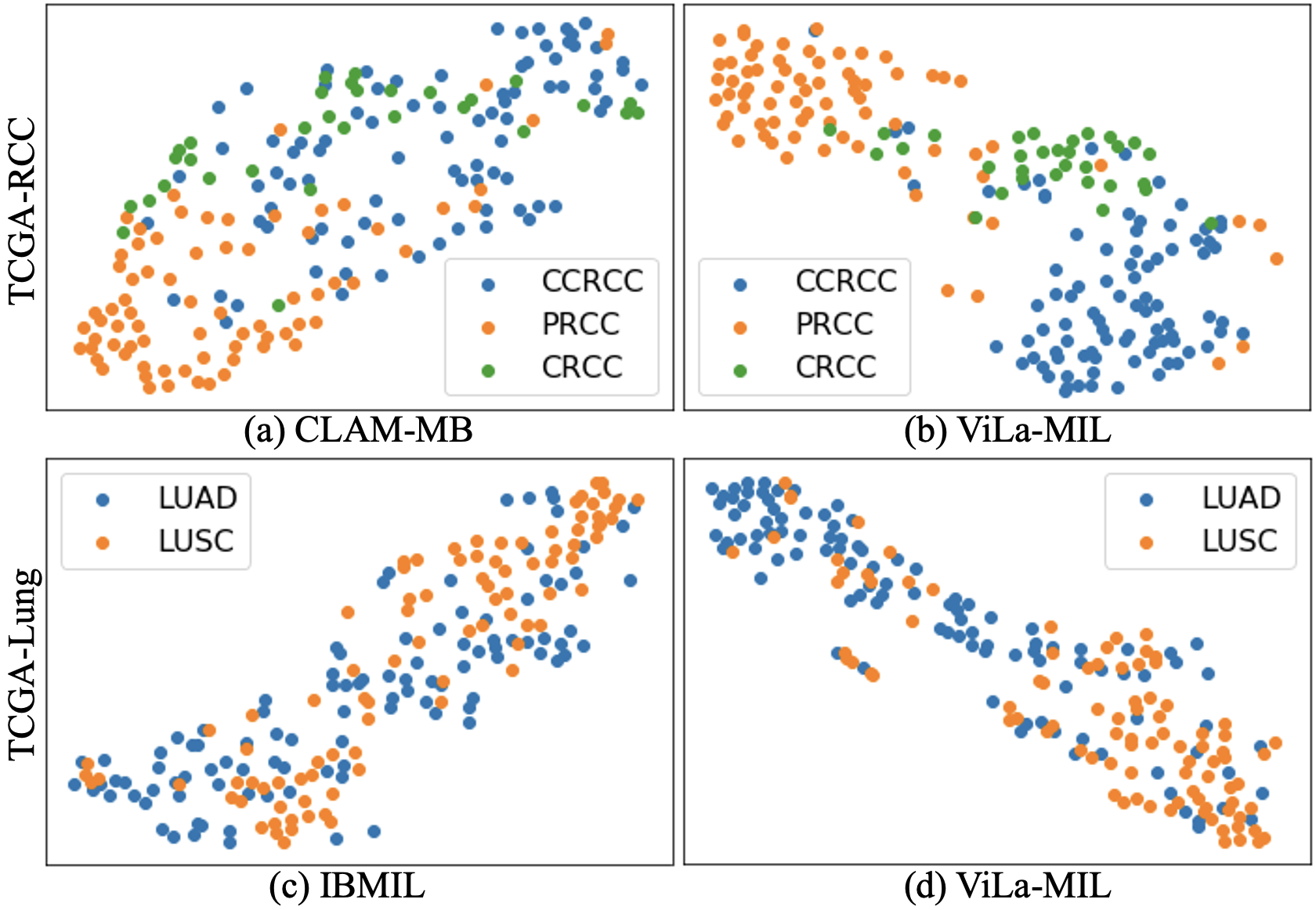}
\caption{Slide-level feature clustering results of different methods on the TCGA-RCC~(top) and TCGA-Lung~(bottom) datasets.
}
\label{clustering_result}
\vspace{-0.6cm}
\end{figure}

\begin{figure*}[h]
\centering
\includegraphics[width=\linewidth]{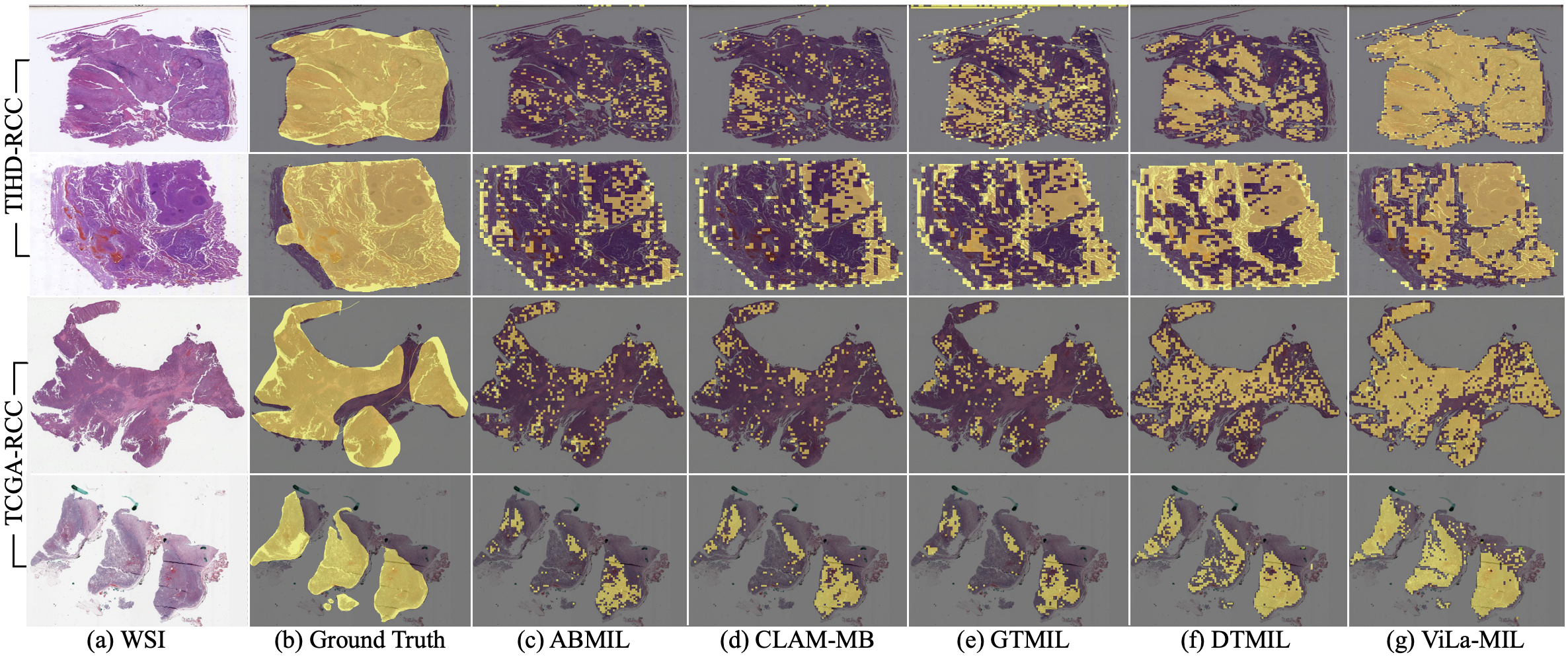}
\caption{
Interpretability analysis~(yellow for cancer) of several exemplars from the TIHD-RCC and TCGA-RCC datasets.
}
\label{prediction_result}
\vspace{-0.5cm}
\end{figure*}

\subsection{Comparisons with State-of-the-Art}
We compare our ViLa-MIL with state-of-the-art pathological image classification methods, including Max-pooling, Mean-pooling, ABMIL \cite{ilse2018attention}, CLAM \cite{lu2021data}, TransMIL \cite{shao2021transmil}, DSMIL \cite{li2021dual}, GTMIL \cite{zheng2021deep}, DTMIL \cite{zhang2022dtfd}, and IBMIL \cite{lin2023interventional}. 
DSMIL \cite{li2021dual} and our ViLa-MIL utilize the patches cropped from 5$\times$ and 10$\times$ slides. The other methods are single-scale approaches and use the patches cropped from 10$\times$ slides. We also compare with two single-scale variants of our method: ViLa-MIL Low, which utilizes 5$\times$ slide and low-scale text prompt, and ViLa-MIL High, which utilizes the 10$\times$ slide and high-scale text prompt.
All the comparison methods utilize the ResNet50-based frozen CLIP image encoder to extract the patch features to ensure a fair comparison.

The experiment results are reported in Table \ref{exp1}. Compared with baselines~(\textit{i.e.}, ABMIL), our ViLa-MIL achieves significant performance improvement across all metrics on all three datasets. Moreover, compared with the current best MIL-based methods, ViLa-MIL achieves an improvement of 1.7-7.2\% in AUC, 2.1-7.3\% in F1, and 0.8-6.6\% in ACC across three datasets.
Traditional MIL-based methods rely on a large number of slides with bag-level labels to train the model. This training paradigm is unable to learn discriminative visual features under the few-shot setting.
In this work, even though the training data is limited, our ViLa-MIL can still learn discriminative morphological patterns from the WSI. 
Specifically, with the guidance of the visual descriptive text prompt, the single-scale ViLa-MILs already demonstrate promising performance, while the performance is boosted further with the dual-scale text prompt and image features.
Moreover, the proposed two decoders transfer the CLIP knowledge efficiently for the WSI subtyping task, which further improves the model's classification performance.

In Figure \ref{clustering_result}, we visualize the slide-level feature clustering results by t-SNE \cite{van2008visualizing} on the validation sets of TCGA-RCC and TCGA-Lung. 
Compared with the best MIL-based methods, ViLa-MIL learns an embedding space that exhibits a higher level of intra-class compactness and inter-class separability, which further demonstrates its superiority. 
In Figure \ref{prediction_result}, we also visualize qualitative prediction results for further comparison.
Compared with several current best methods, ViLa-MIL demonstrates better localization of the cancer region. 

\begin{table}[]
\centering
\scalebox{0.65}{
\begin{tabular}{@{}l|lll|lll@{}}
\toprule
Setting      & \multicolumn{3}{c|}{TIHD $\rightarrow$ TCGA} & \multicolumn{3}{c}{TCGA $\rightarrow$ TIHD} \\ \midrule
Metric & \multicolumn{1}{c}{AUC} & \multicolumn{1}{c}{F1} & \multicolumn{1}{c|}{ACC} & \multicolumn{1}{c}{AUC} & \multicolumn{1}{c}{F1} & \multicolumn{1}{c}{ACC} \\ \midrule
Mean & 61.9$\pm$9.2    & 31.9$\pm$7.7   & 24.6$\pm$3.6   & 69.1$\pm$5.9    & 43.9$\pm$7.1   & 34.0$\pm$6.5   \\
ABMIL        & 61.6$\pm$7.0    & 26.9$\pm$2.7   & 24.7$\pm$3.3   & 66.8$\pm$6.1    & 43.1$\pm$6.9   & 34.1$\pm$8.6   \\
CLAM-MB         & 77.1$\pm$6.9    & 36.1$\pm$2.0   & 35.3$\pm$2.6   & 68.2$\pm$8.7    & 46.9$\pm$8.2   & 35.0$\pm$6.6   \\
TransMIL     & 71.1$\pm$5.9    & 50.3$\pm$6.3   & 47.3$\pm$4.8   & 63.2$\pm$7.9    & 44.6$\pm$7.5   & 30.9$\pm$8.4   \\
DSMIL        & 74.3$\pm$7.3    & 24.6$\pm$4.6   & 21.9$\pm$6.7   & 68.1$\pm$6.0    & 42.1$\pm$6.0   & 32.3$\pm$6.3   \\
GTMIL        & 76.7$\pm$9.2    & 46.7$\pm$4.8   & 43.1$\pm$8.6   & 72.7$\pm$9.7    & 51.0$\pm$7.4$^*$   & 37.4$\pm$5.7$^*$   \\
DTMIL        & 77.9$\pm$5.6    & 30.0$\pm$6.8   & 30.9$\pm$5.7   & 69.8$\pm$8.3    & 38.0$\pm$6.6   & 35.4$\pm$7.0   \\
IBMIL        & 78.2$\pm$4.7    &37.0$\pm$6.5   & 36.9$\pm$4.6   & 70.1$\pm$7.4    & 47.0$\pm$6.4   & 35.5$\pm$7.2   \\
\rowcolor{gray!20} \textbf{ViLa-MIL}  & \textbf{83.4$\pm$3.3}     & \textbf{51.7$\pm$5.6}   & \textbf{50.3$\pm$5.8}    & \textbf{75.4$\pm$4.9}     & \textbf{51.4$\pm$4.7}    & \textbf{38.3$\pm$5.4}     \\ \bottomrule
\end{tabular}}
\caption{Results~(presented in \%) of multi-center cross-evaluation between TIHD-RCC and TCGA-RCC for renal cell carcinoma subtyping under the 16-shot setting. The best result is in bold and its comparable performance is denoted by superscript * based on a paired t-text~(p-value $>$ 0.05).}
\label{exp2}
\vspace{-0.5cm}
\end{table}

\subsection{Generalization on Domain Shift}
To verify the generalization capabilities of ViLa-MIL, we report the cross-evaluation results for renal cell carcinoma between TIHD-RCC and TCGA-RCC. As shown in Table \ref{exp2}, the first two columns denote the results of training on TIHD-RCC and validating on TCGA-RCC; the last two columns show the results of exchanging the training and validation datasets. 
Due to the differences in data distribution, the performance of all methods has decreased. 
ViLa-MIL still achieves an improvement of 5.5\% in the AUC metric compared with the best baseline, indicating its superior cross-domain adaptability and robustness. 
Although the data distribution may vary across different institutions, the diagnostic criteria and visual description features remain constant. 
Our dual-scale visual descriptive text prompt contains comprehensive pathological diagnostic information, ensuring ViLa-MIL to have a better generalization ability across multiple centers.

\subsection{Ablation Studies}
\noindent\textbf{Effect of Each Module in ViLa-MIL.}
To evaluate the effect of each module in ViLa-MIL, we perform an ablation study on the TIHD-RCC dataset under the 16-shot setting. 
For more details on module ablation settings, please refer to Supplementary Material.
The experiment results are reported in Table \ref{exp3}.
Specifically, the first two rows are the baseline results that utilize the single-scale text prompt and attention-based patch aggregation method~(\textit{i.e.}, ABMIL \cite{ilse2018attention}). After replacing the ABMIL \cite{ilse2018attention} with our proposed prototype-guided patch decoder~(\textit{i.e.}, the third and fourth rows), all three metrics have increased, which indicates that our proposed patch decoder is more efficient in aggregating these large number of patch features by grouping similar patch features into the same prototype and progressively generating the final slide features. After introducing our dual-scale text prompt~(\textit{i.e.}, the fifth row), we observe a sensible improvement for all three metrics. This indicates that our dual-scale visual descriptive text prompt can boost the WSI classification performance by effectively utilizing the complementary image features at multiple magnifications. The last row is the results of our ViLa-MIL, by introducing the context-guided text decoder. The model's performance also shows a certain amount of improvement, which denotes that the multi-granular patch and prototype features can refine the text features further. 

\noindent\textbf{Effect of Patch Aggregation.}
To verify the effect of patch aggregation methods, we compare our prototype-guided patch decoder with commonly used feature aggregation methods~(\textit{i.e.}, mean pooling, attention-based pooling, and self-attention-based pooling). 
Specifically, we replace our proposed patch decoder with these three methods, respectively. 
As shown in Table \ref{abl1}, our decoder achieves the best results under all three metrics on the TIHD-RCC dataset, which means that our progressive aggregation method is more effective by grouping similar patch features into the same prototype, then fusing all the prototypes to generate the final slide features.

\begin{table}[]
\label{ablation}
\centering
\scalebox{0.85}{
\begin{tabular}{@{}l|lll@{}}
\toprule
Method                                             & \multicolumn{1}{c}{AUC} & \multicolumn{1}{c}{F1} & \multicolumn{1}{c}{ACC} \\ \midrule
ABMIL + Low-scale     & 76.8 $\pm$ 3.1 & 57.2 $\pm$ 3.9 & 61.4 $\pm$ 2.8 \\
ABMIL + High-scale    & 79.3 $\pm$ 3.2 & 62.7 $\pm$ 3.6 & 63.3 $\pm$ 3.1 \\
Patch Decoder + Low-scale  & 79.4 $\pm$ 2.1 & 60.8 $\pm$ 5.1 & 62.8 $\pm$ 3.9 \\
Patch Decoder + High-scale & 82.9 $\pm$ 2.6 & 64.8 $\pm$ 5.1 & 65.6 $\pm$ 4.7 \\
Patch Decoder + Dual-scale   & 83.6 $\pm$ 2.7              & 67.8 $\pm$ 4.5             & 68.3 $\pm$ 4.1              \\
\rowcolor{gray!20} \textbf{ViLa-MIL} & \textbf{84.3 $\pm$ 4.6}              & \textbf{68.7 $\pm$ 7.3}             & \textbf{68.8 $\pm$ 7.3}              \\ \bottomrule
\end{tabular}}
\caption{Ablation experiment~(presented in \%) on the TIHD-RCC dataset under 16-shot setting.}
\label{exp3}
\vspace{-0.4cm}
\end{table}

\begin{table}[]
\centering
\scalebox{0.85}{
\begin{tabular}{@{}l|ccc@{}}
\toprule
Metric               & AUC        & F1         & ACC        \\ \midrule
Mean Pooling         & 80.8 $\pm$ 4.5 & 66.6 $\pm$ 5.2 & 67.6 $\pm$ 4.9 \\
Attention-based Pooling     & 81.1 $\pm$ 2.9 & 62.9 $\pm$ 7.5 & 64.9 $\pm$ 4.4 \\
Self-attention-based Pooling & 80.1 $\pm$ 3.2 & 62.3 $\pm$ 4.2 & 62.8 $\pm$ 3.9 \\
\rowcolor{gray!20} \textbf{ViLa-MIL} & \textbf{84.3 $\pm$ 4.6} & \textbf{68.7 $\pm$ 7.3} & \textbf{68.8 $\pm$ 7.3} \\ \bottomrule
\end{tabular}}
\caption{Results~(presented in \%) of different patch feature aggregation methods on the TIHD-RCC dataset under 16-shot setting.}
\label{abl1}
\vspace{-0.5cm}
\end{table}

\noindent\textbf{Effect of Text Prompt.}
To verify the effect of different text prompts, we compare three text prompts~(\textit{i.e.}, ``Class-name-replacement", ``Diagnostic Guidelines", and ``Large Language Model"). 
Specifically, the ``Class-name-replacement" just replaces the class name in the text prompt ``A WSI of \{class name\}";
``Diagnostic Guideline" utilizes the guideline of ``WHO Classification of Tumours" for each subtyping, which is also a handcrafted template with clinical prior;
``Large Language Model" adopts the large language model~(\textit{i.e.}, GPT-3.5) to generate a visual descriptive text prompt but does not differentiate between multi-scale visual features. 
As shown in Table \ref{abl2}, compared with the ``Class-name-replacement", the ``Diagnostic Guideline" achieves better performance by introducing more diagnostic-related texts.
Our ViLa-MIL significantly outperforms all the other methods, indicating its capability to learn more discriminative visual features and improve the model's classification ability for challenging and ambiguous samples. 

\noindent\textbf{Effect of Large Language Model.}
To verify the effect of different large language models, we compare our adopted GPT-3.5 with three current popular LLMs~(\textit{i.e.}, PaLM-2 \cite{chowdhery2022palm}, LLaMA-2 \cite{touvron2023llama} and GPT-4 \cite{GPT4}). 
These LLMs utilize the same question prompt to generate the dual-scale visual descriptive text prompt like GPT-3.5. 
The results are reported in Table \ref{LLM}. 
All variants significantly outperform other baselines and superior performances have been achieved with different LLMs, indicating high robustness of ViLa-MIL to different kinds of frozen LLMs.
Compared with GPT-3.5, GPT-4 achieves superior performances on metrics of AUC and ACC. 
This implies that the LLM with enhanced language expression capabilities has the potential to further improve the performance of ViLa-MIL. 

\begin{table}[]
\centering
\scalebox{0.85}{
\begin{tabular}{@{}l|ccc@{}}
\toprule
Method                    & AUC        & F1         & ACC        \\ \midrule
Class-name-replacement  & 66.4 $\pm$ 2.5 & 40.0 $\pm$ 4.4 & 44.3 $\pm$ 3.0 \\
Diagnostic Guideline     & 78.0 $\pm$ 2.8 & 59.5 $\pm$ 2.2 & 61.1 $\pm$ 1.2 \\
Large Language Model   & 80.5 $\pm$ 1.5 & 63.3 $\pm$ 4.5 & 64.6 $\pm$ 4.8 \\
\rowcolor{gray!20} \textbf{ViLa-MIL} & \textbf{84.3 $\pm$ 4.6} & \textbf{68.7 $\pm$ 7.3} & \textbf{68.8 $\pm$ 7.3} \\ \bottomrule
\end{tabular}
}
\caption{Results~(presented in \%) of different text prompts on the TIHD-RCC dataset under 16-shot setting.}
\label{abl2}
\vspace{-0.4cm}
\end{table}

\begin{table}[]
\centering
\scalebox{0.9}{
\begin{tabular}{@{}l|ccc@{}}
\toprule
Method   & AUC   & F1    & ACC      \\ \midrule
PaLM-2 \cite{chowdhery2022palm}   & 83.5 $\pm$ 3.6          & 65.8 $\pm$ 3.6          & 66.3 $\pm$ 3.9          \\
LLaMA-2 \cite{touvron2023llama}   & 81.5 $\pm$ 3.3          & 66.2 $\pm$ 7.3          & 66.8 $\pm$ 3.5          \\
\cellcolor{gray!20}{\textbf{GPT-3.5}} \cite{brown2020language}  & 84.3 $\pm$ 4.6 & \cellcolor{gray!20}{\textbf{68.7 $\pm$ 7.3}} & 68.8 $\pm$ 7.3 \\
\cellcolor{gray!20}{\textbf{GPT-4}} \cite{GPT4}  & \cellcolor{gray!20}{\textbf{85.6 $\pm$ 2.5}} & 68.0 $\pm$ 4.5 & \cellcolor{gray!20}{\textbf{69.2 $\pm$ 4.8}} \\
\bottomrule
\end{tabular}
}
\caption{Results~(presented in \%) of different large language models on the TIHD-RCC dataset under 16-shot setting.}
\label{LLM}
\vspace{-0.6cm}
\end{table}

\noindent\section{Conclusion}
In this work, we proposed a dual-scale vision-language multiple instance learning framework~(ViLa-MIL) for whole slide image classification. 
Specifically, inspired by the diagnostic process of pathologists, we utilized the frozen LLM to generate dual-scale visual descriptive text prompts, which correspond to the hierarchical image contexts in WSIs.
To transfer the VLM to process the WSI efficiently, a prototype-guided patch decoder was proposed to progressively aggregate the patch features. 
A context-guided text decoder was also proposed to refine the text prompt features further by utilizing the multi-granular image contexts. 
Extensive comparative and ablation experiments on three cancer subtyping datasets demonstrated that ViLa-MIL achieved state-of-the-art results for whole slide image classification.
We believe that this work will inspire further research by introducing language prior information to efficiently transfer knowledge from large pre-trained models to the field of pathology.

\section*{Acknowledgements}
This work was partially supported by the National Nature Science Foundation of China (62106191), the Project of China Knowledge Centre for Engineering Science and Technology, and the National Research Foundation, Singapore under its AI Singapore Programme (AISG Award No: AISG2-TC-2021-003).

{\small
\bibliographystyle{ieee_fullname}
\bibliography{egbib}

\begin{thebibliography}{10}\itemsep=-1pt

\bibitem{brown2020language}
Tom Brown, Benjamin Mann, Nick Ryder, Melanie Subbiah, Jared~D Kaplan, Prafulla Dhariwal, Arvind Neelakantan, Pranav Shyam, Girish Sastry, Amanda Askell, et~al.
\newblock Language models are few-shot learners.
\newblock {\em Advances in Neural Information Processing Systems}, 33:1877--1901, 2020.

\bibitem{chan2023histopathology}
Tsai~Hor Chan, Fernando~Julio Cendra, Lan Ma, Guosheng Yin, and Lequan Yu.
\newblock Histopathology whole slide image analysis with heterogeneous graph representation learning.
\newblock In {\em Proceedings of the IEEE/CVF Conference on Computer Vision and Pattern Recognition}, pages 15661--15670, 2023.

\bibitem{chowdhery2022palm}
Aakanksha Chowdhery, Sharan Narang, Jacob Devlin, Maarten Bosma, Gaurav Mishra, Adam Roberts, Paul Barham, Hyung~Won Chung, Charles Sutton, Sebastian Gehrmann, et~al.
\newblock Pa{LM}: Scaling language modeling with {P}athways.
\newblock {\em arXiv preprint arXiv:2204.02311}, 2022.

\bibitem{cui2021artificial}
Miao Cui and David~Y Zhang.
\newblock Artificial intelligence and computational pathology.
\newblock {\em Laboratory Investigation}, 101(4):412--422, 2021.

\bibitem{fey2019fast}
Matthias Fey and Jan~Eric Lenssen.
\newblock Fast graph representation learning with {P}y{T}orch {G}eometric.
\newblock {\em arXiv preprint arXiv:1903.02428}, 2019.

\bibitem{gamper2021multiple}
Jevgenij Gamper and Nasir Rajpoot.
\newblock Multiple instance captioning: Learning representations from histopathology textbooks and articles.
\newblock In {\em Proceedings of the IEEE/CVF Conference on Computer Vision and Pattern Recognition}, pages 16549--16559, 2021.

\bibitem{gao2023clip}
Peng Gao, Shijie Geng, Renrui Zhang, Teli Ma, Rongyao Fang, Yongfeng Zhang, Hongsheng Li, and Yu Qiao.
\newblock C{LIP}-{A}dapter: Better vision-language models with feature adapters.
\newblock {\em International Journal of Computer Vision}, pages 1--15, 2023.

\bibitem{guo2023texts}
Zixian Guo, Bowen Dong, Zhilong Ji, Jinfeng Bai, Yiwen Guo, and Wangmeng Zuo.
\newblock Texts as images in prompt tuning for multi-label image recognition.
\newblock In {\em Proceedings of the IEEE/CVF Conference on Computer Vision and Pattern Recognition}, pages 2808--2817, 2023.

\bibitem{han20232vpt}
Cheng Han, Qifan Wang, Yiming Cui, Zhiwen Cao, Wenguan Wang, Siyuan Qi, and Dongfang Liu.
\newblock E\textsuperscript{2}{VPT}: An effective and efficient approach for visual prompt tuning.
\newblock In {\em Proceedings of the IEEE/CVF International Conference on Computer Vision}, pages 17491--17502, 2023.

\bibitem{H2MIL}
Wentai Hou, Lequan Yu, Chengxuan Lin, Helong Huang, Rongshan Yu, Jing Qin, and Liansheng Wang.
\newblock H$^2$-{MIL}: Exploring hierarchical representation with heterogeneous multiple instance learning for whole slide image analysis.
\newblock In {\em Proceedings of the AAAI Conference on Artificial Intelligence}, pages 933--941, 2022.

\bibitem{huang2023conslide}
Yanyan Huang, Weiqin Zhao, Shujun Wang, Yu Fu, Yuming Jiang, and Lequan Yu.
\newblock Con{S}lide: Asynchronous hierarchical interaction {T}ransformer with breakup-reorganize rehearsal for continual whole slide image analysis.
\newblock In {\em Proceedings of the IEEE/CVF International Conference on Computer Vision}, pages 21349--21360, 2023.

\bibitem{huang2023visual}
Zhi Huang, Federico Bianchi, Mert Yuksekgonul, Thomas~J Montine, and James Zou.
\newblock A visual-language foundation model for pathology image analysis using medical {T}witter.
\newblock {\em Nature Medicine}, pages 1--10, 2023.

\bibitem{ikezogwo2023quilt}
Wisdom Ikezogwo, Saygin Seyfioglu, Fatemeh Ghezloo, Dylan Geva, Fatwir Sheikh~Mohammed, Pavan~Kumar Anand, Ranjay Krishna, and Linda Shapiro.
\newblock Quilt-1{M}: One million image-text pairs for histopathology.
\newblock {\em Advances in Neural Information Processing Systems}, 36, 2024.

\bibitem{ilse2018attention}
Maximilian Ilse, Jakub Tomczak, and Max Welling.
\newblock Attention-based deep multiple instance learning.
\newblock In {\em Proceedings of the International Conference on Machine Learning}, pages 2127--2136, 2018.

\bibitem{khattak2023maple}
Muhammad~Uzair Khattak, Hanoona Rasheed, Muhammad Maaz, Salman Khan, and Fahad~Shahbaz Khan.
\newblock Ma{P}le: Multi-modal prompt learning.
\newblock In {\em Proceedings of the IEEE/CVF Conference on Computer Vision and Pattern Recognition}, pages 19113--19122, 2023.

\bibitem{li2021dual}
Bin Li, Yin Li, and Kevin~W Eliceiri.
\newblock Dual-stream multiple instance learning network for whole slide image classification with self-supervised contrastive learning.
\newblock In {\em Proceedings of the IEEE/CVF Conference on Computer Vision and Pattern Recognition}, pages 14318--14328, 2021.

\bibitem{li2023task}
Honglin Li, Chenglu Zhu, Yunlong Zhang, Yuxuan Sun, Zhongyi Shui, Wenwei Kuang, Sunyi Zheng, and Lin Yang.
\newblock Task-specific fine-tuning via variational information bottleneck for weakly-supervised pathology whole slide image classification.
\newblock In {\em Proceedings of the IEEE/CVF Conference on Computer Vision and Pattern Recognition}, pages 7454--7463, 2023.

\bibitem{li2022blip}
Junnan Li, Dongxu Li, Caiming Xiong, and Steven Hoi.
\newblock B{LIP}: Bootstrapping language-image pre-training for unified vision-language understanding and generation.
\newblock In {\em International Conference on Machine Learning}, pages 12888--12900. PMLR, 2022.

\bibitem{li2023weakly}
Kailu Li, Ziniu Qian, Yingnan Han, I Eric, Chao Chang, Bingzheng Wei, Maode Lai, Jing Liao, Yubo Fan, and Yan Xu.
\newblock Weakly supervised histopathology image segmentation with self-attention.
\newblock {\em Medical Image Analysis}, 86:102791, 2023.

\bibitem{li2023scaling}
Yanghao Li, Haoqi Fan, Ronghang Hu, Christoph Feichtenhofer, and Kaiming He.
\newblock Scaling language-image pre-training via masking.
\newblock In {\em Proceedings of the IEEE/CVF Conference on Computer Vision and Pattern Recognition}, pages 23390--23400, 2023.

\bibitem{lin2023interventional}
Tiancheng Lin, Zhimiao Yu, Hongyu Hu, Yi Xu, and Chang-Wen Chen.
\newblock Interventional bag multi-instance learning on whole-slide pathological images.
\newblock In {\em Proceedings of the IEEE/CVF Conference on Computer Vision and Pattern Recognition}, pages 19830--19839, 2023.

\bibitem{lin2023clip}
Yuqi Lin, Minghao Chen, Wenxiao Wang, Boxi Wu, Ke Li, Binbin Lin, Haifeng Liu, and Xiaofei He.
\newblock C{LIP} is also an efficient segmenter: A text-driven approach for weakly supervised semantic segmentation.
\newblock In {\em Proceedings of the IEEE/CVF Conference on Computer Vision and Pattern Recognition}, pages 15305--15314, 2023.

\bibitem{lu2023towards}
Ming~Y Lu, Bowen Chen, Drew~FK Williamson, Richard~J Chen, Ivy Liang, Tong Ding, Guillaume Jaume, Igor Odintsov, Andrew Zhang, Long~Phi Le, et~al.
\newblock Towards a visual-language foundation model for computational pathology.
\newblock {\em arXiv preprint arXiv:2307.12914}, 2023.

\bibitem{lu2023visual}
Ming~Y Lu, Bowen Chen, Andrew Zhang, Drew~FK Williamson, Richard~J Chen, Tong Ding, Long~Phi Le, Yung-Sung Chuang, and Faisal Mahmood.
\newblock Visual language pretrained multiple instance zero-shot transfer for histopathology images.
\newblock In {\em Proceedings of the IEEE/CVF Conference on Computer Vision and Pattern Recognition}, pages 19764--19775, 2023.

\bibitem{lu2021data}
Ming~Y Lu, Drew~FK Williamson, Tiffany~Y Chen, Richard~J Chen, Matteo Barbieri, and Faisal Mahmood.
\newblock Data-efficient and weakly supervised computational pathology on whole-slide images.
\newblock {\em Nature Biomedical Engineering}, 5(6):555--570, 2021.

\bibitem{maron1997framework}
Oded Maron and Tom{\'a}s Lozano-P{\'e}rez.
\newblock A framework for multiple-instance learning.
\newblock {\em Advances in Neural Information Processing Systems}, 10:570--576, 1997.

\bibitem{nakhli2023sparse}
Ramin Nakhli, Puria~Azadi Moghadam, Haoyang Mi, Hossein Farahani, Alexander Baras, Blake Gilks, and Ali Bashashati.
\newblock Sparse multi-modal graph {T}ransformer with shared-context processing for representation learning of giga-pixel images.
\newblock pages 11547--11557, 2023.

\bibitem{niazi2019digital}
Muhammad Khalid~Khan Niazi, Anil~V Parwani, and Metin~N Gurcan.
\newblock Digital pathology and artificial intelligence.
\newblock {\em The Lancet Oncology}, 20(5):e253--e261, 2019.

\bibitem{GPT4}
OpenAI.
\newblock G{PT}-4 technical report.
\newblock {\em arXiv preprint arXiv:2303.08774}, 2023.

\bibitem{qu2024rise}
Linhao Qu, Kexue Fu, Manning Wang, Zhijian Song, et~al.
\newblock The rise of ai language pathologists: Exploring two-level prompt learning for few-shot weakly-supervised whole slide image classification.
\newblock {\em Advances in Neural Information Processing Systems}, 36, 2024.

\bibitem{radford2021learning}
Alec Radford, Jong~Wook Kim, Chris Hallacy, Aditya Ramesh, Gabriel Goh, Sandhini Agarwal, Girish Sastry, Amanda Askell, Pamela Mishkin, Jack Clark, et~al.
\newblock Learning transferable visual models from natural language supervision.
\newblock In {\em International Conference on Machine Learning}, pages 8748--8763, 2021.

\bibitem{rao2022denseclip}
Yongming Rao, Wenliang Zhao, Guangyi Chen, Yansong Tang, Zheng Zhu, Guan Huang, Jie Zhou, and Jiwen Lu.
\newblock Dense{CLIP}: Language-guided dense prediction with context-aware prompting.
\newblock In {\em Proceedings of the IEEE/CVF Conference on Computer Vision and Pattern Recognition}, pages 18082--18091, 2022.

\bibitem{ryu2023ocelot}
Jeongun Ryu, Aaron~Valero Puche, JaeWoong Shin, Seonwook Park, Biagio Brattoli, Jinhee Lee, Wonkyung Jung, Soo~Ick Cho, Kyunghyun Paeng, Chan-Young Ock, et~al.
\newblock O{CELOT}: Overlapped cell on tissue dataset for histopathology.
\newblock In {\em Proceedings of the IEEE/CVF Conference on Computer Vision and Pattern Recognition}, pages 23902--23912, 2023.

\bibitem{shao2021transmil}
Zhuchen Shao, Hao Bian, Yang Chen, Yifeng Wang, Jian Zhang, Xiangyang Ji, et~al.
\newblock Trans{MIL}: Transformer based correlated multiple instance learning for whole slide image classification.
\newblock {\em Advances in Neural Information Processing Systems}, 34:2136--2147, 2021.

\bibitem{10244116}
Jiangbo Shi, Lufei Tang, Zeyu Gao, Yang Li, Chunbao Wang, Tieliang Gong, Chen Li, and Huazhu Fu.
\newblock M{G}-{T}rans: Multi-scale graph transformer with information bottleneck for whole slide image classification.
\newblock {\em IEEE Transactions on Medical Imaging}, 42(12):3871--3883, 2023.

\bibitem{shi2023structure}
Jiangbo Shi, Lufei Tang, Yang Li, Xianli Zhang, Zeyu Gao, Yefeng Zheng, Chunbao Wang, Tieliang Gong, and Chen Li.
\newblock A structure-aware hierarchical graph-based multiple instance learning framework for p{T} staging in histopathological image.
\newblock {\em IEEE Transactions on Medical Imaging}, 42(10):3000--3011, 2023.

\bibitem{song2023artificial}
Andrew~H Song, Guillaume Jaume, Drew~FK Williamson, Ming~Y Lu, Anurag Vaidya, Tiffany~R Miller, and Faisal Mahmood.
\newblock Artificial intelligence for digital and computational pathology.
\newblock {\em Nature Reviews Bioengineering}, pages 1--20, 2023.

\bibitem{tang2023multiple}
Wenhao Tang, Sheng Huang, Xiaoxian Zhang, Fengtao Zhou, Yi Zhang, and Bo Liu.
\newblock Multiple instance learning framework with masked hard instance mining for whole slide image classification.
\newblock In {\em Proceedings of the IEEE/CVF International Conference on Computer Vision}, pages 4078--4087, 2023.

\bibitem{touvron2023llama}
Hugo Touvron, Thibaut Lavril, Gautier Izacard, Xavier Martinet, Marie-Anne Lachaux, Timoth{\'e}e Lacroix, Baptiste Rozi{\`e}re, Naman Goyal, Eric Hambro, Faisal Azhar, et~al.
\newblock L{L}a{MA}: Open and efficient foundation language models.
\newblock {\em arXiv preprint arXiv:2302.13971}, 2023.

\bibitem{van2008visualizing}
Laurens Van~der Maaten and Geoffrey Hinton.
\newblock Visualizing data using t-{SNE}.
\newblock {\em Journal of Machine Learning Research}, 9(11):2579--2605, 2008.

\bibitem{wu2023cora}
Xiaoshi Wu, Feng Zhu, Rui Zhao, and Hongsheng Li.
\newblock C{ORA}: Adapting {CLIP} for open-vocabulary detection with region prompting and anchor pre-matching.
\newblock In {\em Proceedings of the IEEE/CVF Conference on Computer Vision and Pattern Recognition}, pages 7031--7040, 2023.

\bibitem{xu2019camel}
Gang Xu, Zhigang Song, Zhuo Sun, Calvin Ku, Zhe Yang, Cancheng Liu, Shuhao Wang, Jianpeng Ma, and Wei Xu.
\newblock C{AMEL}: A weakly supervised learning framework for histopathology image segmentation.
\newblock In {\em Proceedings of the IEEE/CVF International Conference on computer vision}, pages 10682--10691, 2019.

\bibitem{yu2023task}
Tao Yu, Zhihe Lu, Xin Jin, Zhibo Chen, and Xinchao Wang.
\newblock Task residual for tuning vision-language models.
\newblock In {\em Proceedings of the IEEE/CVF Conference on Computer Vision and Pattern Recognition}, pages 10899--10909, 2023.

\bibitem{zhang2022dtfd}
Hongrun Zhang, Yanda Meng, Yitian Zhao, Yihong Qiao, Xiaoyun Yang, Sarah~E Coupland, and Yalin Zheng.
\newblock D{TFD-MIL}: Double-tier feature distillation multiple instance learning for histopathology whole slide image classification.
\newblock In {\em Proceedings of the IEEE/CVF Conference on Computer Vision and Pattern Recognition}, pages 18802--18812, 2022.

\bibitem{zhang2023large}
Sheng Zhang, Yanbo Xu, Naoto Usuyama, Jaspreet Bagga, Robert Tinn, Sam Preston, Rajesh Rao, Mu Wei, Naveen Valluri, Cliff Wong, et~al.
\newblock Large-scale domain-specific pretraining for biomedical vision-language processing.
\newblock {\em arXiv preprint arXiv:2303.00915}, 2023.

\bibitem{zhang2023text}
Yunkun Zhang, Jin Gao, Mu Zhou, Xiaosong Wang, Yu Qiao, Shaoting Zhang, and Dequan Wang.
\newblock Text-guided foundation model adaptation for pathological image classification.
\newblock In {\em International Conference on Medical Image Computing and Computer Assisted Intervention}, pages 272--282. Springer, 2023.

\bibitem{zheng2021deep}
Yi Zheng, Rushin~H. Gindra, Emily~J. Green, Eric~J. Burks, Margrit Betke, Jennifer~E. Beane, and Vijaya~B. Kolachalama.
\newblock A graph-{T}ransformer for whole slide image classification.
\newblock {\em IEEE Transactions on Medical Imaging}, 41(11):3003--3015, 2022.

\bibitem{zhong2022regionclip}
Yiwu Zhong, Jianwei Yang, Pengchuan Zhang, Chunyuan Li, Noel Codella, Liunian~Harold Li, Luowei Zhou, Xiyang Dai, Lu Yuan, Yin Li, et~al.
\newblock Region{CLIP}: Region-based language-image pretraining.
\newblock In {\em Proceedings of the IEEE/CVF Conference on Computer Vision and Pattern Recognition}, pages 16793--16803, 2022.

\bibitem{zhou2022learning}
Kaiyang Zhou, Jingkang Yang, Chen~Change Loy, and Ziwei Liu.
\newblock Learning to prompt for vision-language models.
\newblock {\em International Journal of Computer Vision}, 130(9):2337--2348, 2022.

\end{thebibliography}
}

\clearpage

\renewcommand{\thesection}{\Alph{section}}
\renewcommand*{\thefigure}{S\arabic{figure}}
\renewcommand*{\thetable}{S\arabic{table}}

\setcounter{table}{0}
\setcounter{figure}{0}
\setcounter{section}{0}

\begin{table*}[]
\centering
\scalebox{0.9}{
\begin{tabular}{llccc}
\toprule
\multicolumn{2}{l|}{Cancer Type}                         & \multicolumn{2}{c|}{Kidney}               & Lung      \\ \midrule
\multicolumn{2}{l|}{Dataset}                            & TIHD-RCC & \multicolumn{1}{c|}{TCGA-RCC} & TCGA-Lung \\ \midrule
\multicolumn{2}{l|}{Number of WSIs}                      & 480      & \multicolumn{1}{c|}{639}      & 658       \\ \midrule
\multicolumn{1}{l|}{\multirow{3}{*}{Data Split}}        & \multicolumn{1}{l|}{Training} & 192         & \multicolumn{1}{c|}{255}         & 264         \\
\multicolumn{1}{l|}{} & \multicolumn{1}{l|}{Validation} & 144      & \multicolumn{1}{c|}{192}      & 197       \\
\multicolumn{1}{l|}{} & \multicolumn{1}{l|}{Test}       & 144      & \multicolumn{1}{c|}{192}      & 197       \\ \midrule
\multicolumn{1}{l|}{\multirow{2}{*}{Number of Patches}} & \multicolumn{1}{l|}{5$\times$}       & 429,402    & \multicolumn{1}{c|}{607,817}    & 490,977    \\
\multicolumn{1}{l|}{}                                   & \multicolumn{1}{l|}{10$\times$}      & 1,670,362 & \multicolumn{1}{c|}{2,359,471} & 1,903,894 \\ \bottomrule
\end{tabular}}
\caption{Data statistics.}
\label{dataset}
\end{table*}

\noindent\textbf{\Large Supplementary Materials}

\section{Dual-scale Visual Descriptive Text Prompt}
\begin{figure*}
\centering
\includegraphics[width=\linewidth]{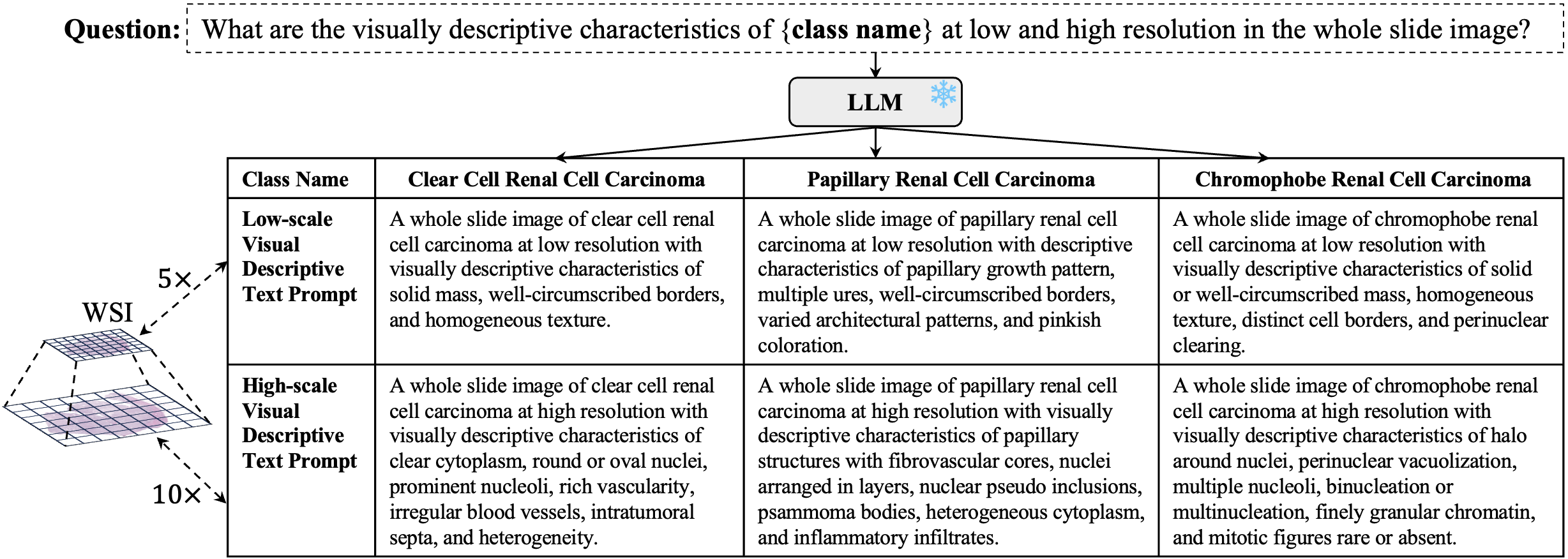}
\caption{Dual-scale visual descriptive text prompt for the renal cell carcinoma. 
By replacing the placeholder ``class name" with the specific category label, such as Clear Cell Renal Cell Carcinoma, Papillary Renal Cell Carcinoma, and Chromophobe Renal Cell Carcinoma, the frozen LLM can generate the dual-scale visual descriptive text prompt that corresponds to the multi-scale WSIs.}
\label{text_prompt}
\end{figure*}

\begin{figure*}
\centering
\includegraphics[width=\linewidth]{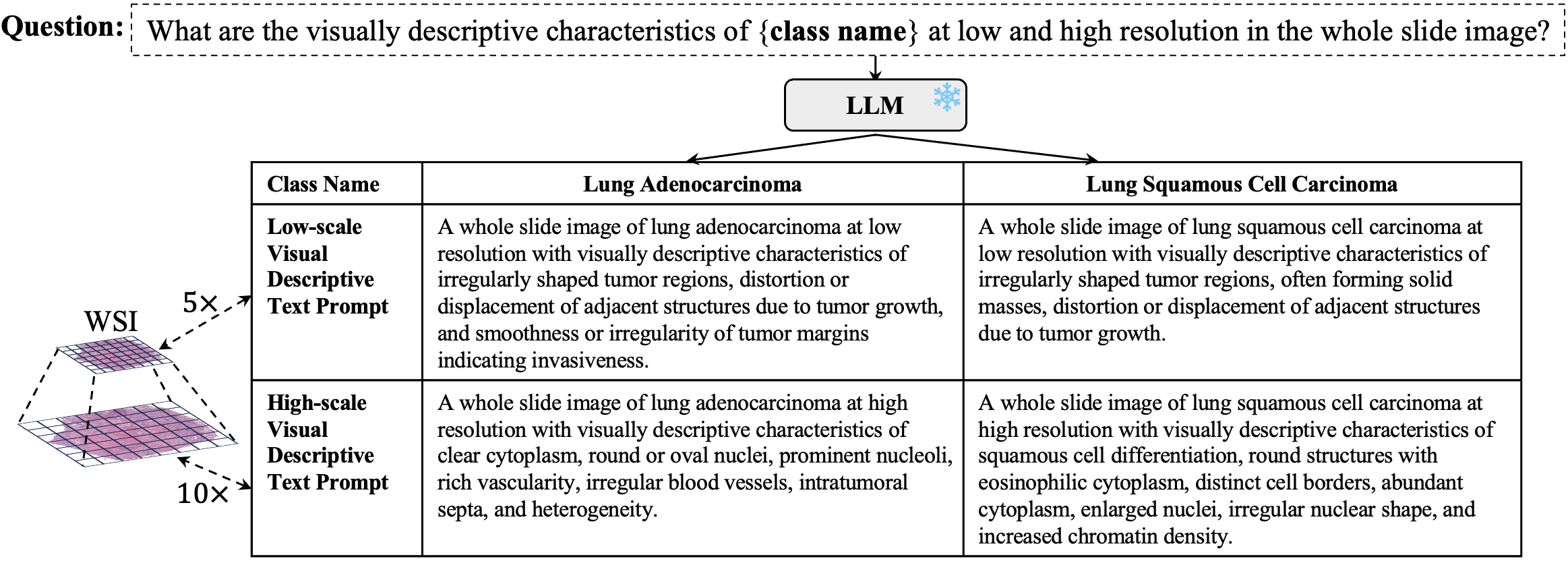}
\caption{Dual-scale visual descriptive text prompt for the lung cancer. 
By replacing the placeholder ``class name" with the specific category label, such as Lung Adenocarcinoma and Lung Squamous Cell Carcinoma, the frozen LLM can generate the dual-scale visual descriptive text prompt that corresponds to the multi-scale WSIs.}
\label{text_prompt_lung}
\end{figure*}

The specific descriptions of dual-scale visual descriptive text prompts for renal cell carcinoma and lung cancer are shown in Figure \ref{text_prompt} and Figure \ref{text_prompt_lung}, respectively. Note that three experienced pathologists thoroughly examined the text prompts generated by GPT-3.5, and found them relatively correct and detailed.

\section{Description of Datasets}
The data statistics are reported in Table \ref{dataset}. The specific descriptions of our collected datasets are as follows:
\begin{itemize}[itemsep=0pt]
    \item \textbf{TIHD-RCC}: This is the in-house dataset consisting of renal cell carcinoma slides collected by our research team. All the slides were stained with hematoxylin and eosin~(H\&E) and scanned by a KF-PRO-005 digital slice scanner at 20 $\times$ magnification with 0.5 $\mu$m/pixel resolution. There are 480 slides with slide-level subtyping labels. 
    \item \textbf{TCGA-RCC}: To verify ViLa-MIL on multi-center data, we collected 639 renal cell carcinoma slides from TCGA.\footnote{https://portal.gdc.cancer.gov.} The data in TIHD-RCC and TCGA-RCC are divided into three categories: clear cell~(CCRCC), papillary~(PRCC), and chromophore renal cell carcinoma~(CRCC).  
    \item \textbf{TCGA-Lung}: To verify the generalizability of ViLa-MIL across multiple cancer types, we also collected 658 lung cancer slides from TCGA. All the slides were annotated with slide-level subtyping labels. This dataset includes two subtypes: lung adenocarcinoma~(LUAD) and lung squamous cell carcinoma~(LUSC). 
\end{itemize}

\section{Implementation Details}
Additional descriptions of implementation details are as follows.
We adopt Adam optimization with a learning rate of 1 $\times$ 10$^{-4}$ and weight decay of 1 $\times$ 10$^{-5}$. The minimum training epoch number is 80, and the early stopping strategy is adopted if the accuracy does not continuously increase for 20 epochs. The batch size is 1. 
During training, all the methods utilize the same seed.
ViLa-MIL and all the baselines are implemented with PyTorch and the PyG library \cite{fey2019fast} on a workstation with eight NVIDIA 2080Ti GPUs. 

\section{Description of Compared Methods}
The specific description for each compared method is as follows:
\begin{itemize}[itemsep=0pt]
    \item \textbf{Max-pooling}: Max-pooling is a baseline method that utilizes the max-pooling operator to generate the slide prediction.
    \item \textbf{Mean-pooling}: Mean-pooling is a baseline method that utilizes the mean-pooling operator to aggregate all the patch features as the slide features.
    \item \textbf{ABMIL \cite{ilse2018attention}}: ABMIL proposes an attention-based aggregation method to generate the slide features by measuring the importance of each instance.
    \item \textbf{CLAM \cite{lu2021data}}: CLAM proposes a global pooling operator trained for weakly-supervised slide-level classification tasks. CLAM-SB and CLAM-MB denote the single-attention-branch and multi-attention-branch versions of CLAM, respectively.
    \item \textbf{TransMIL \cite{shao2021transmil}}: TransMIL proposes to utilize the self-attention mechanism to explore the global relations between patches.
    \item \textbf{DSMIL \cite{li2021dual}}: DSMIL utilizes the multi-scale patches as the input and proposes a non-local attention-based fusion method.
    \item \textbf{GTMIL \cite{zheng2021deep}}: GTMIL employs a graph representation to model the WSI data and utilizes a vision Transformer to generate slide features.
    \item \textbf{DTMIL \cite{zhang2022dtfd}}: DTMIL proposes a double-tier MIL framework by introducing the concept of pseudo-bags.
    \item \textbf{IBMIL \cite{lin2023interventional}}: IBMIL proposes an interventional training method based on the backdoor adjustment.
\end{itemize}

\section{Comparisons with SOTA under Different Shots}
\begin{figure*}
\centering
\includegraphics[width=\linewidth]{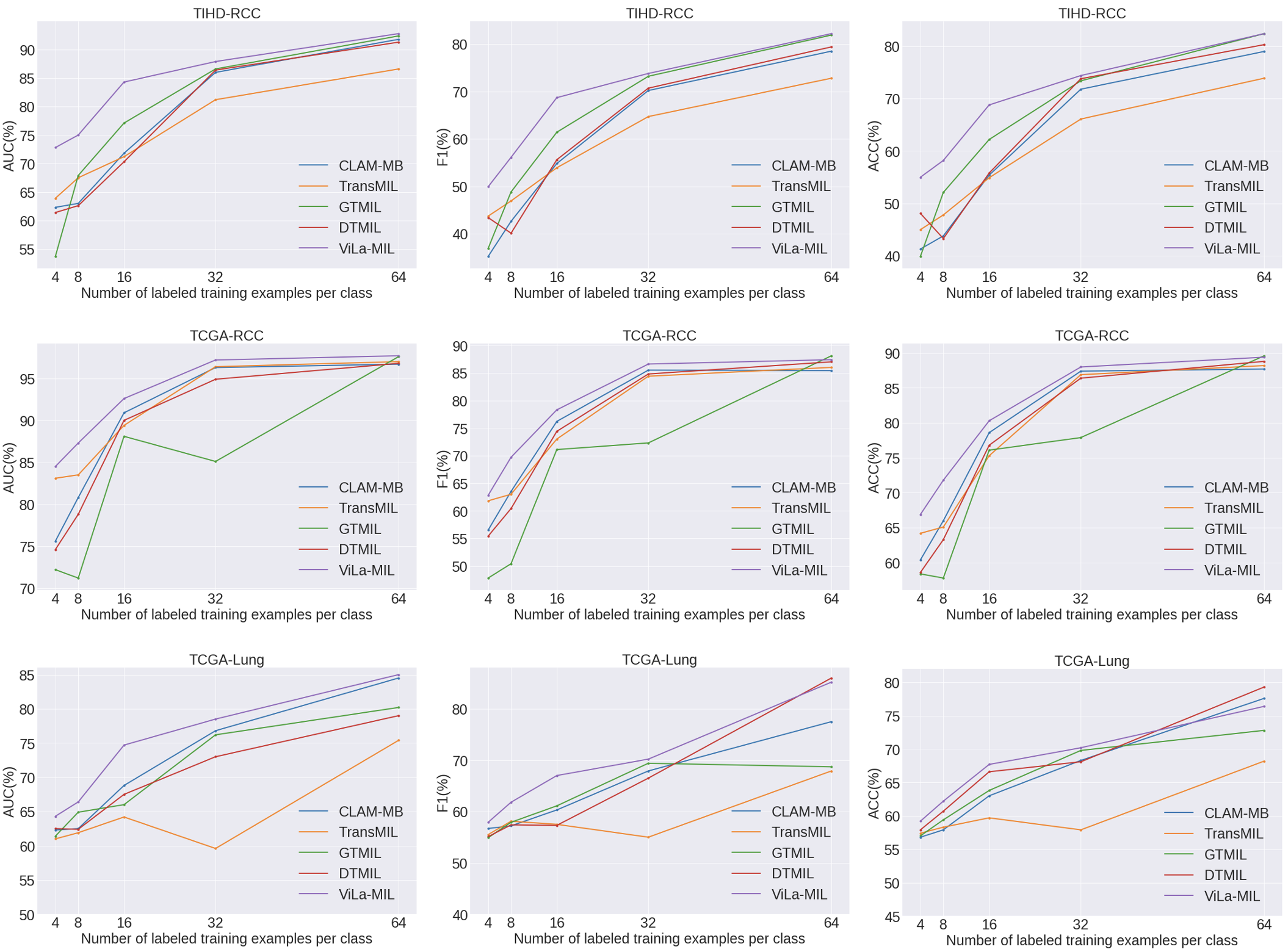}
\caption{Performance comparison with different shots~(4-/8-/16-/32-/64-shot) on TIHD-RCC, TCGA-RCC, and TCGA-lung datasets. N-shot denotes that each class has N training samples.}
\label{different_shots}
\end{figure*}

To compare our ViLa-MIL with the current state-of-the-art WSI classification methods under different shots, we select four MIL-based methods~(\textit{i.e.}, CLAM-MB \cite{lu2021data}, TransMIL \cite{shao2021transmil}, GTMIL \cite{zheng2021deep} and DTMIL \cite{zhang2022dtfd}). The experiment results are summarized in Figure \ref{different_shots}. 
As the number of shots increases, the performance of nearly all methods improves. Specifically, with fewer support samples~(\textit{i.e.}, 4-shot or 8-shot), our ViLa-MIL achieves more significant gains compared with all the other methods. This indicates that with the help of our dual-scale visual descriptive text prompt, ViLa-MIL can capture discriminative morphological patterns better for classification under the few-shot setting. With more support samples~(\textit{i.e.}, 32-shot or 64-shot), the performance of traditional MIL-based methods increases substantially; however, our ViLa-MIL still demonstrates better~(or at least comparable) performance on all three datasets. 
Note that we have additionally conducted the experiment in a fully supervised setting. Specifically, compared to DTMIL, ViLa-MIL still achieves a comparable performance improvement of 0.8\% and 0.7\% in AUC and F1, respectively.

\section{Interpretability Analysis}
This section elaborates on the generation process of visualization results in Figure 5 of the main text.
For the MIL-based models, we generate the visualization results by binarizing the attention maps. 
For our ViLa-MIL, to keep consistency with the other MIL-based methods, we only visualize the high-resolution visualization result.
Specifically, the prototype-guided attention map~(\textit{i.e.}, $Q_hK_h^{\top}/\sqrt{d} \in \mathbb{R}^{N_h \times N_p}$) is first calculated. This map captures the cross-attention between patches and prototypes. 
Each row in the prototype-guided attention map represents a patch, while each column denotes a prototype. 
To establish the relationships between patches and prototypes, we assign each patch to the prototype with the highest cross-attention value. This grouping process ensures that each patch is associated with a specific prototype.
Next, we employ the attention map $A_h$ to select the prototype with the highest attention value. This prototype is deemed the most representative.
Finally, we consider all the patches that are grouped into the cancer prototype as cancer patches, while the remaining patches are taken as normal.
Note that to obtain the prediction result, MIL-based methods need to carefully select a threshold to binarize the attention map. For different WSIs, the best threshold may be different, it is hard to obtain the most optimal prediction result for each case. Our ViLa-MIL avoids this problem by utilizing the belonging relations between patches and prototypes based on the prototype-guided attention map. 

\section{The Setting of Each Module in Ablation Experiment}
The module ablation settings in the ablation experiment are as follows:
\begin{itemize}[itemsep=0pt]
    \item \textbf{ABMIL + Low-scale}: The attention-based method is utilized to aggregate the low-scale patch features as the baseline. Only the 5$\times$ patches and low-scale visual descriptive text prompt are utilized.
    \item \textbf{ABMIL + High-scale}: The attention-based method is utilized to aggregate the high-scale patch features as the baseline. Only the 10$\times$ patches and high-scale visual descriptive text prompt re utilized.
    \item \textbf{Patch Decoder + Low-scale}: Compared with the ``ABMIL + Low-scale", the attention-based patch feature aggregation method is replaced with our proposed prototype-guided patch feature decoder.
    \item \textbf{Patch Decoder + High-scale}:  Compared with the ``ABMIL + High-scale", the attention-based patch feature aggregation method is replaced with our proposed prototype-guided patch feature decoder.
    \item \textbf{Patch Decoder + Multi-scale}: Compared with ``Patch Decoder + Low-scale", the 10$\times$ patches and the high-scale visual descriptive text prompt are also utilized.
    \item \textbf{ViLa-MIL}: This is our complete framework proposed in this work. Compared with ``Patch Decoder + Multi-scale", the context-guided text decoder is also introduced for both scales.
\end{itemize} 

\section{Effect of Text Prompt}
\begin{figure*}
\centering
\includegraphics[width=\linewidth]{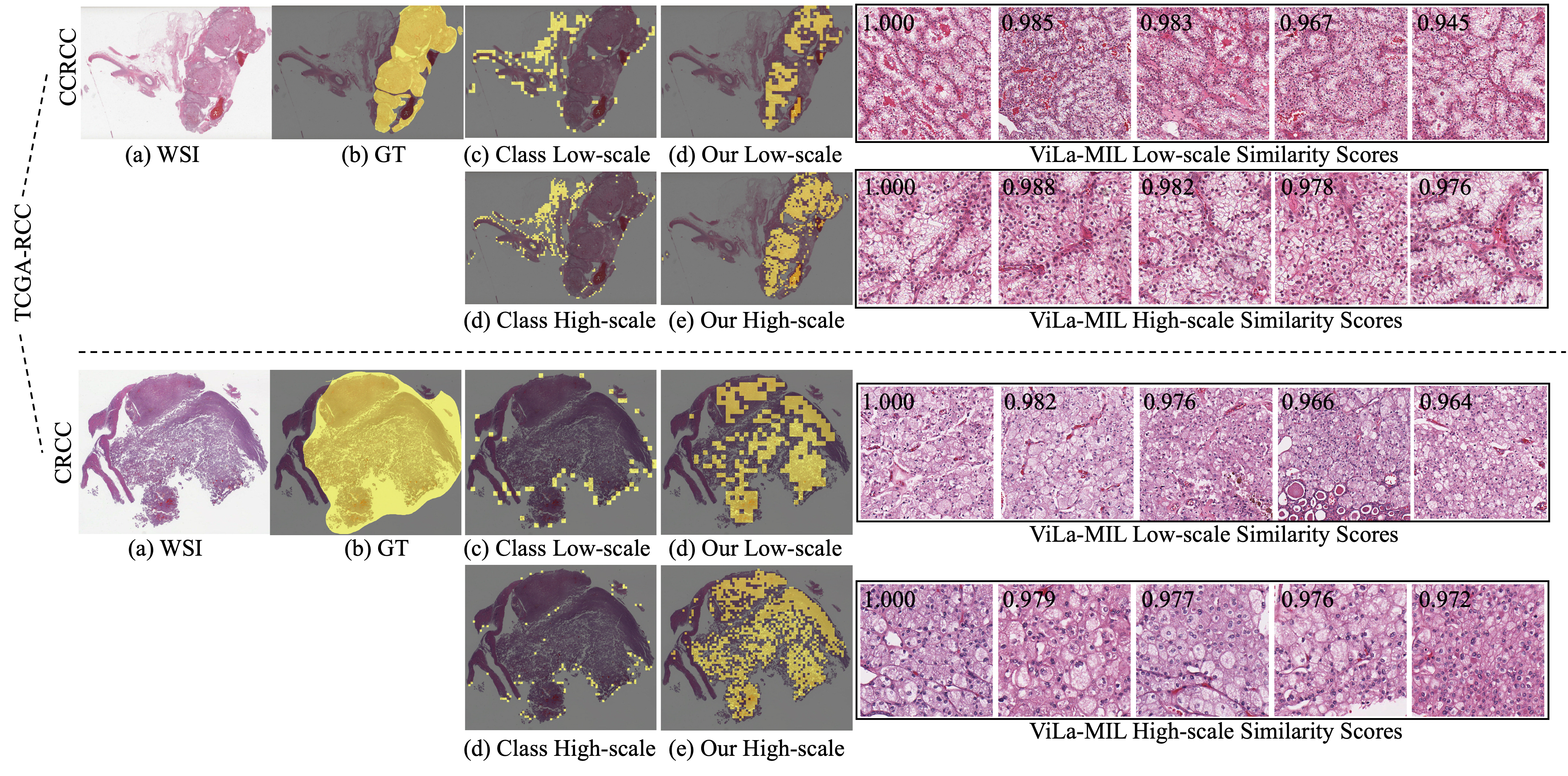}
\caption{Comparison of our dual-scale visual descriptive text prompt with the class-name-replacement text prompt. Two cases~(\textit{i.e.}, CCRCC and CRCC) are randomly selected from the TCGA-RCC dataset to show the results of different text prompts. For each case, (a) is the original WSI; (b) is the corresponding ground truth~(GT) tumor annotation; (c) and (d) are the visualization results by utilizing the ``Class-name-replacement" template at low- and high-scale, respectively; (d) and (e) are the visualization results by utilizing our dual-scale visual descriptive text prompt at low- and high-scale, respectively. 
In the right half of the image, patches with the highest similarity are also visualized at low- and high-scale, respectively.}
\label{different_prompt}
\end{figure*}

To intuitively compare different text prompts, we present the visualization result of the ``Class-name-replacement" template and our ViLa-MIL. As shown in Figure \ref{different_prompt}, compared with the ``Class-name-replacement" template~(the third column), our ViLa-MIL~(the fourth column) achieves better localization results of the tumor. Since the ``Class-name-replacement" template lacks the diagnosis-related prior, it easily fails to locate the tumor regions for subtyping with the supervision of few-shot labels. Our ViLa-MIL maintains a strong consistency with ground truth, which demonstrates that the dual-scale visual descriptive text prompt helps the model learn the discriminative morphological patterns from the WSI. Specifically, as shown in Figures \ref{different_prompt}(d) and \ref{different_prompt}(e), the low-scale visualization result locates the whole tumor structure well, and the high-scale visualization result presents more fine-grained details. 
From the patches with the highest similarity scores to CCRCC, we can observe that the low-scale patches show solid mass, well-circumscribed, and homogeneous texture, and the high-scale patches present clear cytoplasm, prominent nucleoli, and round or oval nuclei. 
Three experienced pathologists also confirm that these highlighted patches capture diagnosis-related patterns for each class. 

\begin{table}[]
\scalebox{0.85}{
\begin{tabular}{@{}l|ccc@{}}
\toprule
Method   & AUC   & F1    & ACC      \\ \midrule
Instance-level + Max Pooling  & 60.4$\pm$6.0          & 39.8$\pm$7.0          & 45.3$\pm$5.1          \\
Instance-level + Top-K    & 74.9$\pm$6.4          & 57.0$\pm$7.6          & 59.4$\pm$7.6          \\
Instance-level + Mean Pooling & 78.0$\pm$2.5          & 60.4$\pm$5.4          & 61.2$\pm$5.3          \\
\rowcolor{gray!20} \textbf{ViLa-MIL}         & \textbf{84.3$\pm$4.6} & \textbf{68.7$\pm$7.3} & \textbf{68.8$\pm$7.3} \\ \bottomrule
\end{tabular}
}
\caption{Results~(presented in \%) of different similarity measurements on the TIHD-RCC dataset under 16-shot setting.}
\label{similarity_calculation}
\end{table}

\section{Effect of Similarity Measurements}
To verify the effect of different similarity measurements, we compare our bag-level similarity with several instance-level methods. 
Specifically, for the instance-level similarity calculation method, based on our ViLa-MIL, each patch feature directly calculates the similarity with the text prompt features, and then several aggregation methods are utilized to obtain the final slide-level similarity. 
As shown in Table \ref{similarity_calculation}, ViLa-MIL achieves the best results under all three metrics because the small patch contains limited visual information, which cannot match various kinds of text descriptions well. 

\section{Effect of Multi-scale Fusion Methods}
\begin{table}[]
\centering
\scalebox{0.9}{
\begin{tabular}{@{}l|ccc@{}}
\toprule
Method   & AUC   & F1    & ACC      \\ \midrule
Feature Summation    & 79.3$\pm$2.9          & 62.7$\pm$4.3          & 65.1$\pm$2.3          \\
\rowcolor{gray!20} \textbf{Logit Summation}         & \textbf{84.3$\pm$4.6} & \textbf{68.7$\pm$7.3} & \textbf{68.8$\pm$7.3} \\ \bottomrule
\end{tabular}
}
\caption{Result~(presented in \%) of different multi-scale fusion methods on the TIHD-RCC dataset under 16-shot setting.}
\label{multi_scale fusion}
\end{table}

To verify the effect of different multi-scale fusion methods, we compare our adopted logit summation with the feature summation method. 
In logit summation, the similarity between image and text features is calculated on each scale separately, and then the  similarities of both scales are added together.
Feature summation means that two-scale image and text features are summed together first, and then the similarity is calculated between the fused image and text features. 
As shown in Table \ref{multi_scale fusion}, the logit summation method achieves better results compared with the feature summation method. 
Due to disparities in visual features at different scales and their corresponding textual descriptions, the feature summation method cannot achieve proper alignment for each scale, resulting in a decrease in performance.

\section{Effect of Hyper-parameters}
\begin{figure*}
\centering
\includegraphics[width=0.85\linewidth]{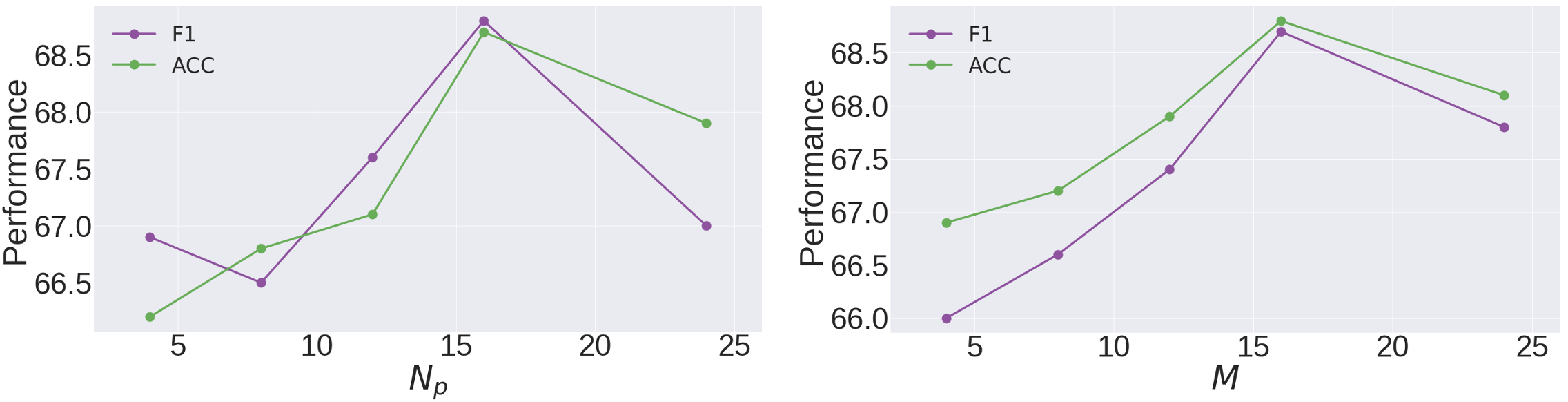}
\caption{Impact of hyper-parameters: the number of prototypes $N_p$~(left) and the number of learnable vectors $M$~(right).}
\label{hyper_params}
\end{figure*}

To verify the effect of hyper-parameters on the model's performance, we conduct a series of ablation studies on the TIHD-RCC dataset. The results are summarized in Figure \ref{hyper_params}. The best results are achieved with 16 prototypes~(\textit{i.e.}, $N_p = 16$). Too few prototypes cannot represent various kinds of visual features sufficiently. On the other hand, with too many prototypes, the redundant prototypes do not contribute to improving the model's performance but increase computational complexity. 
For the number of learnable text vectors $M$ in the text prompt, the best value is 16. Fewer text vectors cannot effectively transfer the model to the pathological dataset. Conversely, employing more text vectors increases the number of parameters the model needs to learn. In the few-shot scenario, optimizing the model based on these limited data becomes challenging.

\section{Effect of CLIP as the backbone}
To verify the effect of domain-related VLMs on the model's performance, we replaced CLIP with PLIP and QuiltNet and conducted the experiments on the TIHD-RCC dataset in the 16-shot setting. Specifically, compared with CLIP (84.3\%), PLIP (85.7\%) \cite{huang2023visual} and QuiltNet (85.2\%) \cite{ikezogwo2023quilt} exhibit AUC improvements of 1.4\% and 0.9 \%. This indicates that VLMs pre-trained on the domain-specific data contribute to enhancing model performance further.


\clearpage

\end{document}